\documentclass[10pt,twocolumn,letterpaper]{article}
\pdfoutput=1

\usepackage[pagenumbers]{cvpr}

\usepackage{multirow}
\usepackage{colortbl}
\usepackage{pifont}
\usepackage{algorithm}
\usepackage{algpseudocode}
\usepackage{bm}

\definecolor{oursblue}{HTML}{E6F3FF}
\definecolor{avggray}{HTML}{F2F2F2}
\definecolor{ourslavender}{HTML}{F0F0FF}

\newcommand{\oursrowS}{\rowcolor{ourslavender}}
\newcommand{\oursrowE}{\rowcolor{oursblue}}
\newcolumntype{g}{>{\columncolor{avggray}}c}

\AtBeginEnvironment{table}{}
\AtBeginEnvironment{table*}{}

\newcommand{\model}{DeLoVi\xspace}
\newcommand{\modelS}{DeLoVi-S\xspace}
\newcommand{\modelE}{DeLoVi-E\xspace}

\definecolor{modelScolor}{HTML}{D55E00}
\definecolor{modelEcolor}{HTML}{0072B2}

\DeclareCaptionTextFormat{colormodels}{\let\modelS\modelScolored\let\modelE\modelEcolored #1}
\AtEndPreamble{%
  \crefname{table}{Table}{Tables}%
  \crefname{figure}{Figure}{Figures}%
}

\definecolor{cvprblue}{rgb}{0.21,0.49,0.74}
\ifdefined\supponly
  \usepackage{xr-hyper}
  \makeatletter
  \def\@newl@bel#1#2#3{\global\@namedef{#1@#2}{#3}}
  \makeatother
\fi
\usepackage[pagebackref,breaklinks,colorlinks,allcolors=cvprblue]{hyperref}

\def\confName{CVPR}
\def\confYear{2027}

\title{Searching Ultra-Long Videos with Decision Models}

\author{
Ce Zhang\quad
Baiqi Li\quad
Seongsu Ha\quad
Gedas Bertasius\\
University of North Carolina at Chapel Hill
}

\begin{document}
\ifdefined\supponly
\setcounter{figure}{\getrefnumber{main:lastfigure}}
\setcounter{table}{\getrefnumber{main:lasttable}}
\else
\maketitle

\begin{abstract}
Understanding ultra-long videos requires retrieving sparse, query-relevant evidence from extensive video histories.
Existing similarity-based retrieval methods are efficient, but they encode the question and the video content independently and thus often miss evidence whose relevance requires reasoning over both.
Generative large language models (LLMs) can perform such reasoning, but they typically generate text over multiple rounds of search, which makes each query slow and costly.
In this work, we introduce \model (\textbf{De}cision Models for \textbf{Lo}ng \textbf{Vi}deo), a framework that leverages pretrained decision models to retrieve evidence from ultra-long videos.
Like an LLM, a decision model reads the question together with the candidate evidence, so it can reason over both.
Unlike an LLM, it directly outputs a relevance score without generating text, which makes it fast and inexpensive.
We first present \modelS, a simple method that generates captions for short video clips and uses a decision model to rank those captions by their relevance to a question.
A generative LLM then answers the question using the top-ranked captions.
Despite its simplicity, \modelS achieves state-of-the-art performance on three ultra-long video understanding benchmarks, EgoLifeQA, EgoMemReason, and MM-Lifelong, and generalizes effectively to long-video understanding benchmarks, VideoMME-L and LVBench.
However, \modelS scores every caption for every question, so its cost grows linearly with video length.
To reduce this cost, we propose an automated policy evolution framework that iteratively refines the memory construction and retrieval policies around a fixed pretrained decision model.
The resulting method, \modelE, maintains comparable performance while reducing input tokens by $34\times$ on average compared with \modelS.
It also runs over $3\times$ faster than the best prior method on EgoLifeQA while achieving higher accuracy.
\iftoggle{cvprfinal}{Our code is available at \url{https://github.com/CeeZh/DeLoVi}.}{}
\end{abstract}

\vspace{-0.5cm}
\section{Introduction}
\label{sec:intro}

\begin{figure}[t]
    \centering
    \includegraphics[width=\linewidth]{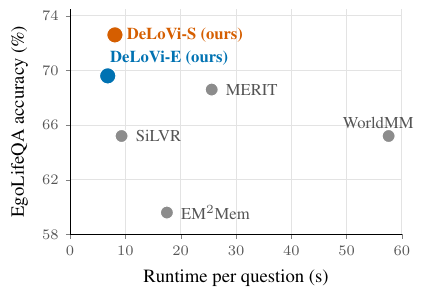}
    \vspace{-0.6cm}
    \caption{\textbf{Accuracy and runtime on EgoLifeQA.} All methods use GPT-6 Luna as the LLM, and runtime is measured per question.
    \modelS achieves the highest accuracy, and \modelE is the fastest method while achieving the second-highest accuracy.
    Compared with MERIT~\cite{choi2026merit}, the strongest prior method, \modelE is $3.8\times$ faster with $1.0\%$ higher accuracy.\vspace{-0.45cm}}
    \label{fig:teaser}
\end{figure}

Understanding ultra-long videos that span days, weeks, or even months is essential for personal AI assistants. Consider a user who asks, ``Where did I leave my spare key after I used it last week?'' Answering this question requires locating when the user last used the key and identifying where they put it afterward. Although the relevant evidence may occupy only a few seconds of video, the assistant must find it among days of unrelated activity.

Recent approaches address this challenge by converting the video into an external memory, typically textual captions or summaries of short clips, and retrieving the memory entries needed to answer a question.
These methods commonly find the entries with similarity-based retrieval, which ranks them by keyword overlap or embedding similarity with the question~\cite{che2026lvmh, luo2024videorag, yang2025egolife}, or with generative large language models (LLMs), which read the entries and decide which ones are relevant~\cite{tian2025egor1, yeo2026worldmm, choi2026merit}.
Similarity-based retrieval is efficient, but it compares the question with each entry only through separately computed encodings, so it often misses evidence whose relevance requires reasoning about how the two relate.
For instance, when a user asks where they left their spare key, many moments in the video involve a key, but only a few show where it was put away.
Telling these moments apart requires understanding what happens in each of them, which is difficult to capture with a similarity score alone.
Generative LLMs can perform this reasoning, but they are slow and expensive at this scale.
The captions of a week-long video alone amount to more than one million tokens, and those of a month-long video to several million, which is far too much for an LLM to process for every question.

Recent advances in pretrained decision models~\cite{typesafe2026jev, perplexity2026decider, dhankhar2026drex, duan2026general} provide a promising alternative to both similarity-based retrieval and generative LLMs.
Given a question and a set of candidate choices, a decision model directly outputs a relevance score or a selection probability for each candidate.
Prior work has demonstrated the effectiveness of decision models on classification and semantic judgment~\cite{deusser2026evaluating, duan2026general}, agent control~\cite{zhu2026jevagentharness}, memory management~\cite{jiang2026jevmem}, and document retrieval~\cite{wan2026decisionrag}.
However, their potential for retrieving evidence from ultra-long videos remains largely unexplored.
We argue that decision models are well suited to this task, which requires judging the relevance of a very large number of candidates.
First, a decision model is fast and cheap, because it produces its judgment without generating any text.
This makes it practical to judge thousands of candidates for a single question, which would be far more expensive with a generative LLM.
Second, unlike similarity-based retrieval, a decision model processes the question and each candidate jointly, so it can judge whether the candidate helps answer the question rather than how closely it matches the question.

Motivated by these properties, we present \model (\textbf{De}cision models for \textbf{Lo}ng \textbf{Vi}deo), a framework that leverages pretrained decision models to retrieve evidence from ultra-long videos.
We first introduce \modelS, which uses a decision model in the simplest possible way.
\modelS divides a video into short clips, describes each clip with a textual caption, and asks the decision model to score every caption independently for its relevance to the question.
The highest-scoring captions are then given to a generative LLM, which answers the question.
Despite its simplicity, \modelS achieves state-of-the-art results on three ultra-long video understanding benchmarks, EgoLifeQA~\cite{yang2025egolife}, EgoMemReason~\cite{wang2026egomemreason}, and MM-Lifelong~\cite{chen2026mmlifelong}.
On EgoLifeQA, it outperforms MERIT~\cite{choi2026merit}, the strongest prior method, by $4.0\%$ (Figure~\ref{fig:teaser}).
It also generalizes effectively to hour-long videos on VideoMME-L~\cite{fu2025videomme} and LVBench~\cite{wang2025lvbench}.

However, \modelS still scores every caption for every question.
On the week-long videos of EgoLifeQA, this amounts to over one million input tokens per question, or more than $500$ million tokens for its $500$ questions, and the cost grows linearly with the length of the video.
To reduce this cost, we introduce \modelE, which organizes the captions into a memory and uses the same decision model to search this memory instead of scoring every caption.
Rather than designing the memory and the search procedure by hand, we obtain them with an automated policy evolution framework, which represents the memory construction and retrieval policies as executable programs.
We initialize the programs with a hierarchy of temporal summaries and a recursive search.
A coding agent then iteratively revises both programs on a set of development questions, using feedback from retrieval trajectories and downstream question answering, with the decision model kept fixed.
After the evolution, the programs are fixed and applied to all benchmarks without further changes.
On EgoLifeQA, the resulting \modelE uses $29\times$ fewer input tokens per question than \modelS.
As shown in Figure~\ref{fig:teaser}, it is also over $3\times$ faster than MERIT while achieving higher accuracy.
Taken together, these results show that a pretrained decision model, used without any model training, is both more accurate and more efficient than prior state-of-the-art methods at retrieving evidence from ultra-long videos.

\section{Related Work}
\label{sec:related}

\paragraph{Memory and retrieval for ultra-long video understanding.}
Language-based approaches, such as LLoVi~\cite{zhang2024llovi} and SiLVR~\cite{zhang2026silvr}, represent videos as textual descriptions for long-range reasoning.
Recent methods explore more structured video memories, including graph-based knowledge~\cite{ren2026videorag}, hierarchical episodic memories~\cite{islam2024videorecap, lin2026hippomm}, event-centric multimodal memories~\cite{chen2026em2mem}, and persistent episodic and semantic memories~\cite{long2026m3agent}.
Other approaches develop hierarchical video representations and adaptive search mechanisms, including VideoTree~\cite{wang2025videotree}, WorldMM~\cite{yeo2026worldmm}, and VideoARM~\cite{yin2026videoarm}.
To select evidence from such memories, one line of work relies on similarity-based retrieval, such as EgoRAG~\cite{yang2025egolife}, which employs retrieval-augmented question answering over long egocentric histories.
Another line of work explores more flexible, iterative retrieval through generative LLMs.
VideoAgent~\cite{wang2024videoagent}, DVD~\cite{zhang2025dvd}, Ego-R1~\cite{tian2025egor1}, and VTS~\cite{zhang2026vts} employ agentic exploration to progressively acquire relevant video evidence.
MERIT~\cite{choi2026merit} and EgoCITE~\cite{zhang2026egocite} iteratively refine semantic and temporal retrieval queries over multi-key or context-augmented memory indices.
While these approaches demonstrate the effectiveness of memory-based and agentic retrieval, they select evidence with similarity-based matching or generative models.
In contrast, we select evidence with a pretrained decision model, and we show that it achieves state-of-the-art results when applied directly to the captions of short video clips, without building a structured memory.

\paragraph{Decision models.}
Recent advances in pretrained decision models, such as Jev~\cite{typesafe2026jev}, Decider~\cite{perplexity2026decider}, and Drex~\cite{dhankhar2026drex}, enable direct scoring and selection among candidate options without autoregressive generation.
Prior work evaluates their capabilities across classification, semantic judgment, and sequential decision-making~\cite{deusser2026evaluating, duan2026general}, and other studies explore their applications to agent control~\cite{zhu2026jevagentharness} and memory management~\cite{jiang2026jevmem}.
When used for retrieval, a decision model plays a role similar to cross-encoder and LLM-based rerankers~\cite{nogueira2019passage, sun2023rankgpt}, which also judge a query and a candidate jointly, and to reranker-guided search~\cite{macavaney2022adaptive, rgs2025}, which uses such judgments to explore a corpus beyond an initial candidate list.
More closely related, DecisionRAG~\cite{wan2026decisionrag} uses pretrained decision models to navigate hierarchical document collections, and VideoJev~\cite{zhu2026videojev} develops video-specific decision models for temporal grounding and question answering.
In contrast to DecisionRAG, which navigates a document hierarchy that is fixed in advance, we target videos that span days to months and evolve the memory and the search policy automatically.
We also use a general-purpose decision model without the video-specific training of VideoJev.

\paragraph{Automated optimization of agentic systems.}
Recent work has explored automatically improving AI agents and their underlying workflows through iterative feedback.
Automated Design of Agentic Systems (ADAS)~\cite{hu2025adas} and AFlow~\cite{zhang2025aflow} formulate agent design as a search problem over executable programs and workflows.
G\"{o}del Agent~\cite{yin2025godel} and the Darwin G\"{o}del Machine~\cite{zhang2025dgm} investigate recursive self-improvement through agent modifications.
AlphaEvolve~\cite{novikov2025alphaevolve} and Meta-Harness~\cite{lee2026metaharness} demonstrate feedback-driven program optimization.
Recent and concurrent work extends these ideas to long-video understanding~\cite{cui2026metavideoagent, chen2026videoemem, lin2026videoharness}.
In particular, LVMH~\cite{che2026lvmh} formulates the design of long-video memory policies as a multi-objective code search under a fixed answer model, and VideoEvolve~\cite{xu2026videoevolve} uses a memory evolver and a retrieval evolver to adapt what is stored and how it is retrieved.
Our policy evolution builds on this line of work, but we study policies built around a pretrained decision model.
We keep the decision model fixed and evolve the memory construction and retrieval programs around it, with the goal of reducing the cost of exhaustive decision-model retrieval while retaining most of its accuracy.

\section{Method}
\label{sec:method}

\begin{figure*}[t]
    \centering
    \includegraphics[width=\linewidth]{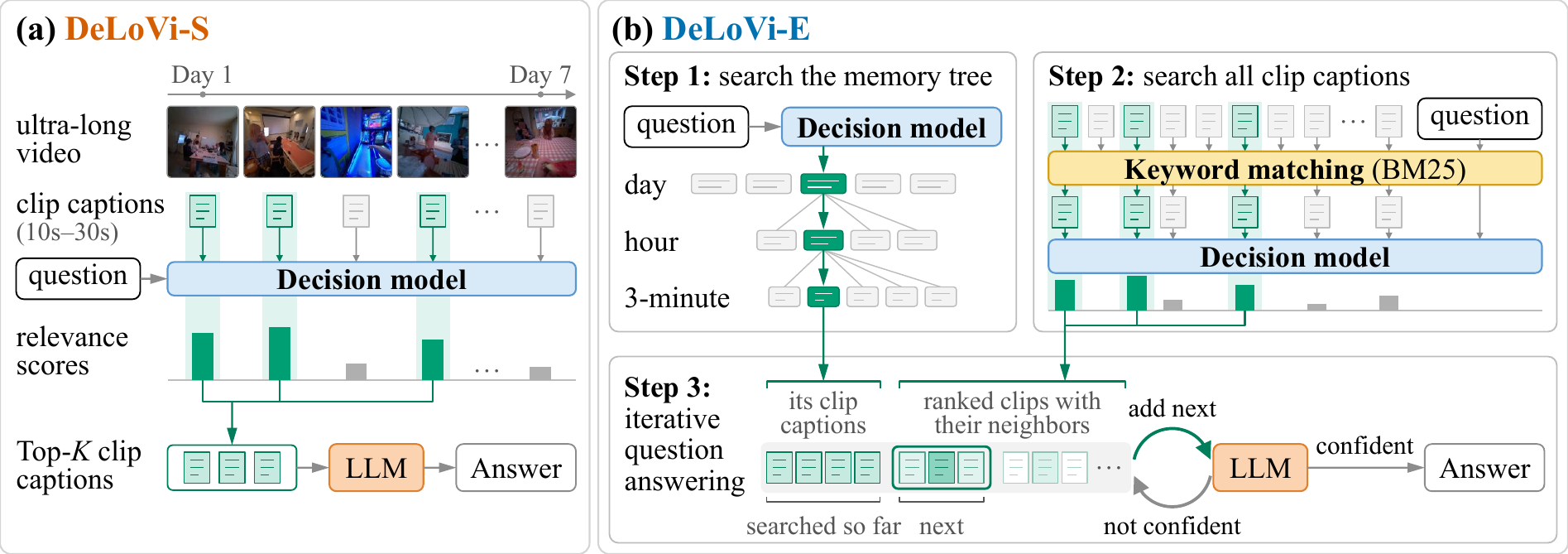}
    \caption{\textbf{Overview of \model.}
    \textbf{(a)} \modelS divides the video into short clips of $10$ to $30$ seconds and captions every clip. A decision model scores the relevance of every clip caption to the question, and an LLM answers the question from the top-$K$ clip captions.
    \textbf{(b)} \modelE answers the question in three steps. In Step 1, the decision model searches over a memory tree and selects the most relevant $3$-minute window. The clip captions of this window are the first input of Step 3. In Step 2, keyword matching (BM25~\cite{robertson2009bm25}) shortlists the clip captions that share words with the question, and the decision model ranks the shortlisted captions by their relevance. Each top-ranked clip is passed to Step 3 together with its neighboring clips. In Step 3, the LLM answers the question from the clips of Step 1 and reports its confidence. If the LLM is not confident, the ranked clips of Step 2 are added in order and the LLM answers again, until it is confident. 
    }
    \label{fig:method}
\end{figure*}

We present \model, a framework that uses a pretrained decision model to retrieve evidence from ultra-long videos (\cref{fig:method}).
After the preliminaries (\cref{sec:formulation}), we introduce \modelS, which uses the decision model to score every clip caption of the video against the question (\cref{sec:delovi_s}).
We then introduce \modelE, which reduces the cost of \modelS with memory and retrieval policies obtained by automated policy evolution (\cref{sec:delovi_e}).

\subsection{Preliminaries}
\label{sec:formulation}

\noindent\textbf{Problem setup.}
Given an ultra-long video $V$ and a question $Q$ asked at time $t$, the goal is to predict the answer $A$ from the video observed before $t$, denoted as $V_{<t}$.
Since $V$ can span days or even months, processing $V_{<t}$ directly with a multimodal LLM is prohibitively expensive.
The main challenge is therefore to find the small part of the video that contains the evidence needed to answer the question.

\noindent\textbf{Decision models.}
A decision model $\mathcal{D}$~\cite{typesafe2026jev, perplexity2026decider, dhankhar2026drex, duan2026general} takes a question $Q$ and a candidate $x$, \eg, the caption of a video clip, and outputs a relevance score $\mathcal{D}(Q, x)$.
A higher score means that the candidate is more useful for the question.
A decision model reads the question and the candidate jointly, as an LLM does.
However, it outputs the score directly instead of generating text, which makes it much faster and cheaper than a generative LLM.
We use a pretrained decision model throughout this work and without training.

\subsection{\modelS: Simple Decision-Model Retrieval}
\label{sec:delovi_s}

Following prior memory-based approaches~\cite{yang2025egolife, yeo2026worldmm, choi2026merit}, \modelS converts the video into a textual memory, retrieves evidence from the memory, and answers the question from the evidence (\cref{fig:method}a).
It implements each of these three stages in the simplest possible way.

For memory construction, we divide the video into short clips (\eg, $10$ to $30$ seconds).
For each clip, we generate a visual caption with a vision-language model and transcribe its speech.
The caption and the transcript together form one memory entry $m_j$.
The memory $\mathcal{M}_t = \{m_j\}$ contains the entries of all clips observed before $t$, in temporal order.

For retrieval, the decision model $\mathcal{D}$ scores every memory entry independently:
\begin{equation}
    s_j = \mathcal{D}(Q, m_j).
    \label{eq:delovi_s}
\end{equation}
We keep the $K$ entries with the highest scores as the evidence set $\mathcal{E}$:
\begin{equation}
    \mathcal{E} = \operatorname*{arg\,top\text{-}K}_{m_j \in \mathcal{M}_t} \; s_j.
    \label{eq:topk}
\end{equation}

For answering, a generative LLM $G$ reads the $K$ retrieved entries in temporal order, each with its timestamp, and predicts the answer $A = G(Q, \mathcal{E})$.

\subsection{\modelE: Evolved Decision-Model Retrieval}
\label{sec:delovi_e}
\label{sec:evolve}

For every question, \modelS scores the caption of every short clip in the video, so its cost grows linearly with the length of the video.
On the week-long videos of EgoLifeQA, \modelS already uses over $1$M input tokens per question (\cref{tab:efficiency}).

To reduce this cost, we formulate the design of a more efficient method as a policy search problem.
A policy consists of a memory policy $F$, which converts the video into a memory, and a retrieval policy $R$, which selects the evidence from the memory with the decision model.
\modelS is one such policy, in which $F$ captions every clip and $R$ scores every caption.
Our goal is to find a policy that is much cheaper than \modelS and remains accurate.
We measure the cost of a policy by the number of input tokens that the decision model and the LLM process per question.

\noindent\textbf{Automated policy evolution.}
Following recent work that optimizes agent harnesses as code~\cite{lee2026metaharness, che2026lvmh}, we implement $F$ and $R$ as executable programs and let a coding agent rewrite them over multiple iterations.
The visual captioner, the decision model, and the LLM stay fixed.
The evolution starts from a hand-designed policy that follows prior tree-based search methods~\cite{zhang2026vts, wang2025videotree, wan2026decisionrag}.
This policy organizes the video into a hierarchy of temporal summaries, and the decision model searches this hierarchy recursively.
We evaluate every policy on a set of development questions by its accuracy and its cost.
We discard a policy as soon as another policy is both more accurate and cheaper.
If one policy is more accurate and another is cheaper, both are kept.
The kept policies form the Pareto frontier of accuracy and cost.
Each iteration starts from one policy on this frontier, which we call the parent policy.
The coding agent inspects the programs of the parent policy, its accuracy and cost, and its retrieval trajectories, which record how each development question was searched and answered.
The agent then rewrites both programs to address the weaknesses it identifies, \eg, questions for which the search missed the relevant clips or used more tokens than needed.
The rewritten programs form a new policy, which is evaluated in the same way.
After the last iteration, every policy on the frontier is the cheapest policy found at its level of accuracy.
Among these policies, we select the one with the highest accuracy on a held-out validation set, and we refer to it as \modelE.
\modelE is then fixed and applied to all evaluation benchmarks without further changes.
Neither the development nor the validation questions overlap with the questions of our evaluation benchmarks (\cref{sec:impl}).
We provide the initial policy and further details in the supplementary material.

\begin{table*}[t]
\centering
\caption{\textbf{Comparison with the state of the art} on ultra-long and long video understanding benchmarks. The LLM of each method is given in parentheses. We report \model with three LLMs, so that it can be compared with each baseline under the same LLM. \modelE uses $34\times$ fewer input tokens than \modelS on average. Best results are in \textbf{bold} and second best are \underline{underlined}.}
\vspace{-0.2cm}

\label{tab:main}
\small
\setlength{\tabcolsep}{6pt}
\begin{tabular}{lccccc}
\toprule
& \multicolumn{3}{c}{\textit{Ultra-long Video}} & \multicolumn{2}{c}{\textit{Long Video}} \\
\cmidrule(lr){2-4} \cmidrule(lr){5-6}
Method & EgoLifeQA & EgoMemReason & MM-Lifelong & VideoMME-L & LVBench \\
\midrule
\multicolumn{6}{l}{\textit{MLLMs Uniform Sampling}} \\
Qwen3-VL-8B~\cite{bai2025qwen3vl}          & 38.6 & 29.6 & --   & 61.0 & 48.3 \\
Gemini 2.5 Pro~\cite{comanici2025gemini25} & 46.4 & --   & --   & 55.7 & 57.0 \\
GPT-5~\cite{openai2025gpt5}                & 48.6 & 27.8 & --   & 74.3 & 60.4 \\
\midrule
\multicolumn{6}{l}{\textit{Similarity-based Methods}} \\
LightRAG~\cite{guo2024lightrag} (GPT-5)    & 48.8 & 26.8   & 15.2   & 46.6 & 30.4 \\
EgoRAG~\cite{yang2025egolife} (GPT-5)      & 52.0 & --   & --   & 41.1 & 32.2 \\
HippoMM~\cite{lin2026hippomm} (GPT-5)      & 54.6 & --   & --   & 41.6 & 38.2 \\
Video-RAG~\cite{luo2024videorag} (GPT-5)   & 55.4 & 32.6   & 22.1   & 55.4 & 33.1 \\
HippoRAG~\cite{gutierrez2024hipporag} (GPT-5) & 59.6 & 27.4   & 22.2   & 52.1 & 54.0 \\
\midrule
\multicolumn{6}{l}{\textit{LLM-based Methods}} \\
Ego-R1~\cite{tian2025egor1} (Qwen2.5-3B)   & 53.0 & 25.8 & --   & 42.7 & 34.1 \\
M3-Agent~\cite{long2026m3agent} (Qwen3-32B) & 53.5 & --   & --   & 55.3 & 49.3 \\
EGAgent~\cite{rege2026egagent} (Gemini 2.5 Pro) & 57.5 & --   & --   & 74.1 & --   \\
SiLVR~\cite{zhang2026silvr} (GPT-6 Luna)  & 65.2 & 39.6 & 36.1 & 78.7 & 69.6 \\
WorldMM~\cite{yeo2026worldmm} (GPT-5)      & 65.6 & 30.6 & 13.8 & 76.6 & 61.9 \\
EM$^2$Mem~\cite{chen2026em2mem} (GPT-5)    & 66.0 & --   & --   & 76.8 & --   \\
MERIT~\cite{choi2026merit} (GPT-5)         & 71.2 & 38.6 & 29.2 & 77.7 & 71.8 \\
\midrule
\oursrowS
\multicolumn{6}{l}{\textit{Ours: best accuracy}} \\
\oursrowS
\textbf{\modelS} (GPT-6 Luna)              & 72.6 & 36.4 & 36.9 & 74.6 & 66.8 \\
\oursrowS
\textbf{\modelS} (GPT-5)                   & \underline{75.2} & 37.2 & 37.1 & 77.8 & 70.1 \\
\oursrowS
\textbf{\modelS} (GPT-6.1 Sol)             & \textbf{77.6} & \textbf{44.4} & \textbf{46.8} & \textbf{87.0} & \underline{78.2} \\
\oursrowE
\multicolumn{6}{l}{\textit{Ours: efficient} ($34\times$ fewer input tokens than \modelS)} \\
\oursrowE
\textbf{\modelE} (GPT-6 Luna)              & 69.6 & 37.6 & 34.5 & 77.2 & 68.1 \\
\oursrowE
\textbf{\modelE} (GPT-5)                   & 72.8 & 37.0 & 33.7 & 78.7 & 70.8 \\
\oursrowE
\textbf{\modelE} (GPT-6.1 Sol)            & 73.4 & \underline{42.2} & \underline{43.6} & \underline{85.3} & \textbf{78.4} \\
\bottomrule
\end{tabular}
\end{table*}

\noindent\textbf{The evolved policy.}
We now describe \modelE, the policy that results from this evolution (\cref{fig:method}b).
Its memory policy organizes the memory entries of \modelS into a tree with four levels, which are short clips, $3$-minute windows, hours, and days.
Each node above the clip level stores an LLM summary of its children, and the tree is built online as the video is observed.

Its retrieval policy first tries to answer the question from a single part of the video and searches the rest of the video only if the LLM is not confident in this answer.
In the first attempt, the decision model descends the tree from the top level to the $3$-minute level.
The clips of the selected $3$-minute window form the first \emph{evidence window}, from which the LLM answers the question.
\modelE prompts the LLM to report a confidence score in $[0, 1]$ together with each answer.
If the confidence exceeds $0.8$, the answer is returned and the search stops.
Otherwise, \modelE searches the entire video for further evidence windows, which helps both when the first attempt selected a wrong part of the video and when the question requires evidence from several parts of the video.
The clips of the video are first shortlisted with BM25~\cite{robertson2009bm25}, a standard keyword-matching function that ranks the clip captions by the words they share with the question.
The decision model then re-ranks the shortlisted clips.
BM25 only narrows down the candidates at a negligible cost, and the decision model still decides their ranking.
Each re-ranked clip forms a further evidence window together with its neighboring clips, which provide the temporal context around the clip.
The evidence windows are added to the evidence in ranked order, and the LLM answers the question again each time their number doubles, \ie, after $2$, $4$, and $8$ windows, and so on.
Since the LLM does not answer after every window, the number of LLM calls grows only logarithmically with the number of evidence windows.
The search ends once the confidence of an answer exceeds $0.8$ or after a maximum of $W$ windows.
We provide the values of $W$ and the full algorithm in the supplementary material.

Overall, the decision model scores only the summaries along a single descent of the tree and a shortlist of clips, instead of every memory entry.
As a result, \modelE uses $29\times$ fewer input tokens per question than \modelS on EgoLifeQA (\cref{tab:efficiency}).

\subsection{Implementation Details}
\label{sec:impl}
For EgoLifeQA and EgoMemReason, we use the official captions and transcripts provided by EgoLife~\cite{yang2025egolife} following prior works~\cite{yeo2026worldmm, choi2026merit}. For other benchmarks, we use GPT-6 Luna~\cite{openai2026gpt6luna} to caption each clip and Whisper~\cite{radford2023whisper} to transcribe its speech. We set the clip length to $10$ seconds for VideoMME-L and LVBench and $30$ seconds for MM-Lifelong.
By default, we use Jev~\cite{typesafe2026jev} as the decision model and GPT-6 Luna as the LLM with default reasoning effort.
For \modelS, we set $K = 100$.
For the evolution framework, we use Claude Code~\cite{anthropic2025claudecode} with Claude Opus 5.5~\cite{anthropic2026opus55} with high reasoning effort as the coding agent.
We run the evolution for $30$ iterations.
The development and validation sets contain $500$ and $350$ questions, respectively, sampled from seven datasets.
Their questions do not overlap with the questions of our evaluation benchmarks (\cref{sec:setup}).
We provide more details in the supplementary material.

\section{Experiments}
\label{sec:experiments}

\begin{table*}[t]
\centering
\caption{\textbf{Efficiency comparison} on EgoLifeQA. All methods use the same LLM (GPT-6 Luna). Search counts all decision-model and LLM calls made to retrieve the evidence, and QA is the final LLM call that answers the question. Best results are in \textbf{bold}.}
\vspace{-0.2cm}

\label{tab:efficiency}
\small
\setlength{\tabcolsep}{5pt}
\begin{tabular}{lccgccgccgc}
\toprule
& \multicolumn{3}{c}{Input Tokens / Q} & \multicolumn{3}{c}{API Cost (\textcent{} / Q)} & \multicolumn{3}{c}{Runtime (s / Q)} & \\
\cmidrule(lr){2-4} \cmidrule(lr){5-7} \cmidrule(lr){8-10}
Method & Search & QA & Total & Search & QA & Total & Search & QA & Total & Accuracy \\
\midrule
EM$^2$Mem~\cite{chen2026em2mem} & 11.5K & 8.6K & \textbf{20.1K} & 0.16 & 0.09 & 0.25 & 14.3 & 3.2 & 17.5 & 59.6 \\
WorldMM~\cite{yeo2026worldmm} & 68.4K & 8.3K & 76.7K & 0.86 & 0.09 & 0.95 & 54.2 & 3.4 & 57.6 & 65.2 \\
SiLVR~\cite{zhang2026silvr} & \textbf{0} & 404.9K & 404.9K & \textbf{0} & 7.60 & 7.60 & \textbf{0} & 9.3 & 9.3 & 65.2 \\
MERIT~\cite{choi2026merit}  & 29.4K & 30.2K  & 59.6K  & 0.39 & 0.31 & 0.70 & 20.6 & 5.0 & 25.6 & 68.6 \\
\oursrowS
\modelS (ours)              & 1.03M & 17.7K  & 1.05M  & 4.49 & 0.18 & 4.67 & 5.2 & 2.9 & 8.1 & \textbf{72.6} \\
\oursrowE
\modelE (ours)             & 32.2K & \textbf{4.0K} & 36.2K & 0.16 & \textbf{0.05} & \textbf{0.21} & 4.4 & \textbf{2.4} & \textbf{6.8} & 69.6 \\
\bottomrule
\end{tabular}
\end{table*}

\subsection{Experimental Setup}
\label{sec:setup}

\noindent\textbf{Benchmarks and metrics.}
We evaluate \model on five benchmarks.
EgoLifeQA~\cite{yang2025egolife} and EgoMemReason~\cite{wang2026egomemreason} contain week-long egocentric videos, and MM-Lifelong~\cite{chen2026mmlifelong} contains $181.1$ hours of video at the day, week, and month scales.
We also evaluate on the long split of Video-MME~\cite{fu2025videomme} (VideoMME-L) and on LVBench~\cite{wang2025lvbench}, which contain hour-long videos.
We report the question-answering accuracy on all benchmarks.
For efficiency, we report the cost of retrieving the evidence and answering the question.
We do not count the cost of memory construction, since the memory is built once as the video is observed and is shared by all questions.

\noindent\textbf{Baselines.}
We compare \model with three groups of methods, which differ in how they select the evidence (\cref{tab:main}).
Multimodal LLMs (MLLMs) process uniformly sampled frames.
Similarity-based methods encode the question and the video content separately and retrieve the evidence by the similarity between the two.
LLM-based methods rely on a generative LLM, which searches the video iteratively (\eg, WorldMM~\cite{yeo2026worldmm} and MERIT~\cite{choi2026merit}), selects the evidence from retrieved candidates (\eg, EM$^2$Mem~\cite{chen2026em2mem}), or processes the captions of the entire video (\eg, SiLVR~\cite{zhang2026silvr}).
In contrast, \model selects the evidence with a decision model.
Since the baselines use different LLMs, we list the LLM of each method and report \model with three LLMs.
For \modelE, we run the retrieval with GPT-6 Luna and only replace the LLM that produces the final answer.

\subsection{Main Results}
\label{sec:main_results}

In \cref{tab:main}, we compare \modelS and \modelE with the three groups of baselines.
Since the accuracy of every method depends on its LLM, we first compare methods that use the same LLM.
First, with GPT-5, the LLM of most baselines, \modelS achieves the best accuracy among all GPT-5-based methods on EgoLifeQA and MM-Lifelong.
It outperforms MERIT, the strongest baseline, by $4.0\%$ and $7.9\%$ on these two benchmarks and is within $1.7\%$ of MERIT on the other three.
It also outperforms the best similarity-based method on every benchmark, by $4.6\%$ to $22.4\%$.
Second, \modelE stays close to \modelS while using $34\times$ fewer input tokens on average.
With GPT-5, it is within $3.4\%$ of \modelS on every benchmark and outperforms MERIT on EgoLifeQA, MM-Lifelong, and VideoMME-L.
On average over the five benchmarks, \modelS and \modelE outperform MERIT by $1.8\%$ and $0.9\%$, respectively.
\modelE is also $3.8\times$ faster and $3.3\times$ cheaper than MERIT on EgoLifeQA when both use the same LLM (\cref{tab:efficiency}).
Third, with GPT-6 Luna, \modelS outperforms SiLVR by $7.4\%$ on EgoLifeQA and $0.8\%$ on MM-Lifelong, while SiLVR is $2.8\%$ to $4.1\%$ higher on the other three benchmarks.
However, SiLVR has the highest API cost among all methods in \cref{tab:efficiency}, because its LLM processes the captions of the entire video for every question.
On EgoLifeQA, SiLVR costs $1.6\times$ more per question than \modelS and $36\times$ more than \modelE.
Lastly, the accuracy of \model improves further with a stronger LLM.
With GPT-6.1 Sol, \modelS achieves the highest accuracy in \cref{tab:main} on four of the five benchmarks.

\subsection{Analysis across Time Scales}
\label{sec:timescale}
\begin{figure}[t]
\centering
\includegraphics[width=\linewidth]{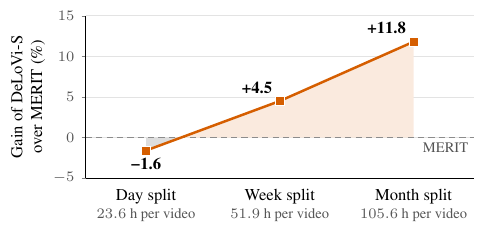}
\vspace{-0.6cm}
\caption{\textbf{Accuracy gain of \modelS over MERIT at different time scales} on MM-Lifelong. To answer a question, each method has to search an entire video, and the average video duration is $23.6$, $51.9$, and $105.6$ hours on the day, week, and month splits, respectively. Both methods use the same LLM (GPT-6 Luna). MERIT is more accurate on the shortest video, whereas \modelS outperforms it on the longer videos.}
\label{fig:timescale}
\vspace{-0.3cm}
\end{figure}

\begin{table*}[t]
\centering
\caption{\textbf{Comparison between the initial policy and the evolved policy} on the five test benchmarks. Both policies use the same LLM (GPT-6 Luna). We report the accuracy ($\%$) and the input tokens per question, averaged over the five benchmarks. Search counts all decision-model and LLM calls to retrieve the evidence, and QA covers the final LLM call that answers the question. \modelE uses $5.6\times$ fewer input tokens than the initial policy with a comparable average accuracy.}
\vspace{-0.2cm}
\label{tab:policy_comparison}
\small
\setlength{\tabcolsep}{3pt}
\begin{tabular}{lccgcccccg}
\toprule
& \multicolumn{3}{c}{Input Tokens / Q} & \multicolumn{6}{c}{Accuracy} \\
\cmidrule(lr){2-4} \cmidrule(lr){5-10}
Method & Search & QA & Total & EgoLifeQA & EgoMemReason & MM-Lifelong & VideoMME-L & LVBench & Avg. \\
\midrule
Initial policy          & 207.1K         & 7.4K          & 214.5K         & 66.2          & \textbf{40.4} & 33.0          & 76.9          & \textbf{68.1} & 56.9          \\
Evolved policy (\modelE) & \textbf{33.9K} & \textbf{4.7K} & \textbf{38.6K} & \textbf{69.6} & 37.6          & \textbf{34.5} & \textbf{77.2} & \textbf{68.1} & \textbf{57.4} \\
\bottomrule
\end{tabular}
\end{table*}

In \cref{fig:timescale}, we plot the accuracy gain of \modelS over MERIT on the day, week, and month splits of MM-Lifelong, using the same LLM (GPT-6 Luna) for both methods.
To answer a question, each method has to search an entire video, and the average video duration is $23.6$, $51.9$, and $105.6$ hours on the day, week, and month splits, respectively.
From the figure, we can observe that the gain of \modelS over MERIT grows with the video duration.
On the shortest video (the day split), MERIT outperforms \modelS by $1.6\%$ ($25.6\%$ vs.\ $24.0\%$).
However, this advantage is reversed once the video is longer.
\modelS outperforms MERIT by $4.5\%$ on the week split ($42.1\%$ vs.\ $37.6\%$), and the gain grows to $11.8\%$ on the month split ($39.3\%$ vs.\ $27.5\%$), where the video is $4.5\times$ longer than on the day split.
These results show that the longer the video, the larger the benefit of searching with a decision model, which makes \model particularly effective for ultra-long videos.

\subsection{Efficiency Analysis}
\label{sec:efficiency}
\begin{figure}[t]
\centering
\includegraphics[width=\linewidth]{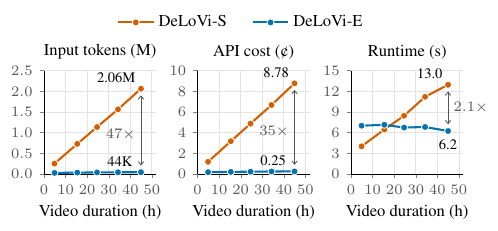}
\vspace{-0.6cm}
\caption{\textbf{Efficiency against video duration} on EgoLifeQA. We report the input tokens, API cost (in cents) and runtime per question. The cost of \modelS grows linearly with the video duration, whereas the cost of \modelE increases only slightly.}
\label{fig:duration}
\vspace{-0.3cm}
\end{figure}

\addtocounter{table}{2}
\begin{table*}[t]
\centering
\providecommand{\nclip}[1]{\,{\scriptsize\color{gray}(#1)}}
\caption{\textbf{Ablation on the number of retrieved captions} $K$ in \modelS. All methods use the same LLM (GPT-6 Luna). MERIT selects the number of clips adaptively, and the gray numbers give the average number of clips that it passes to the LLM per question.}
\vspace{-0.2cm}
\label{tab:abl_num_captions}
\small
\setlength{\tabcolsep}{5pt}
\begin{tabular}{llcccccg}
\toprule
Method & \# Captions & EgoLifeQA & EgoMemReason & MM-Lifelong & VideoMME-L & LVBench & Avg. \\
\midrule
MERIT~\cite{choi2026merit} & 106 & 68.6\nclip{98} & \textbf{39.2}\nclip{167} & 29.1\nclip{154} & 73.2\nclip{47} & \textbf{68.6}\nclip{64} & 55.7\nclip{106} \\
\midrule
\multirow{3}{*}{\modelS (ours)} & 10  & 66.6 & 30.2 & 31.0 & 71.6 & 62.9 & 52.5 \\
                         & 50  & 71.0 & 35.6 & \textbf{37.1} & \textbf{74.9} & 67.1 & 57.1 \\
                         & 100 & \textbf{72.6} & 36.4 & 36.9 & 74.6 & 66.8 & \textbf{57.4} \\
\bottomrule
\end{tabular}
\end{table*}

\addtocounter{table}{-3}

\noindent\textbf{Comparison with prior methods.}
In \cref{tab:efficiency}, we compare the efficiency of \model with four strong LLM-based methods on EgoLifeQA, using the same LLM (GPT-6 Luna) for all methods.
First, we note that both \modelS and \modelE are more accurate than all four baselines, and \modelS outperforms MERIT, the strongest baseline, by $4.0\%$.
Second, \modelE has the lowest API cost and the shortest runtime among all methods.
Compared with \modelS, it uses $29\times$ fewer input tokens and costs $22\times$ less, with $3.0\%$ lower accuracy.
Compared with MERIT, it is $3.8\times$ faster and $3.3\times$ cheaper while achieving $1.0\%$ higher accuracy.
Third, the speed of \modelE comes from searching with a decision model.
\modelE and MERIT use a similar number of input tokens during the search ($32.2$K vs.\ $29.4$K), but the search takes $4.4$ seconds for \modelE and $20.6$ seconds for MERIT, which searches iteratively with a generative LLM.
\modelE also calls the LLM only $1.9$ times per question on average and answers $86.6\%$ of the questions with at most two calls.
Lastly, compared with SiLVR, which gives the captions of the entire video to the LLM, \modelE uses $11\times$ fewer input tokens and costs $36\times$ less while achieving $4.4\%$ higher accuracy.

\begin{figure}[t]
\centering
\includegraphics[width=\linewidth]{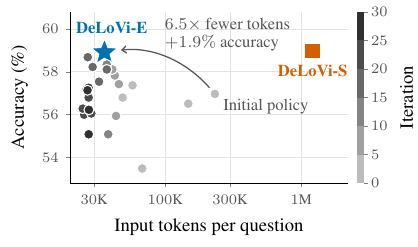}
\vspace{-0.8cm}
\caption{\textbf{Evolution dynamics} on the validation set. Each point is the policy of one iteration, and darker points are later iterations. The input tokens per question (log scale) decrease throughout the evolution, while the accuracy stays close to that of the initial policy. \modelE (star) uses $6.5\times$ fewer tokens than the initial policy with $1.9\%$ higher accuracy. It also achieves similar accuracy to \modelS (square) with $34\times$ fewer tokens.}
\vspace{-0.4cm}
\label{fig:evolution}
\end{figure}

\noindent\textbf{Scaling with video duration.}
In \cref{fig:duration}, we group the questions of EgoLifeQA by the duration of the video observed before the question, which ranges from $0.2$ to $51.9$ hours.
The cost of \modelS grows linearly with the video duration.
From videos shorter than $10$ hours to videos longer than $40$ hours, its input tokens increase from $0.25$M to $2.06$M and its API cost from $1.18$ to $8.78$ cents per question.
In contrast, the input tokens of \modelE only increase from $26.7$K to $44.0$K and its API cost from $0.18$ to $0.25$ cents.
As a result, the token reduction relative to \modelS grows with the video duration, from $9\times$ on the shortest videos to $47\times$ on the longest ones.

\subsection{Evolution Analysis}
\label{sec:evolution}

\noindent\textbf{Evolution dynamics.}
In \cref{fig:evolution}, we show the policies produced in the different iterations of the evolution, evaluated on the validation set.
First, the cost decreases throughout the evolution.
The input tokens drop from $232.2$K for the initial policy to $40.9$K at iteration $10$ and converge to around $27$K in the last iterations.
Second, the accuracy does not degrade.
Except for one policy at iteration $2$ ($53.5\%$), it stays between $55.1\%$ and $58.9\%$, compared with $57.0\%$ for the initial policy.
Third, the selected policy, \modelE, comes from iteration $17$ and has the highest validation accuracy among all iterations.
It is $1.9\%$ more accurate than the initial policy while using $6.5\times$ fewer tokens.
Compared with \modelS, it achieves similar accuracy ($58.9\%$ vs.\ $59.0\%$) with $34\times$ fewer tokens.

\begin{table}[t]
\centering
\caption{\textbf{Ablations on the evolution framework.} The default \modelE is initialized from recursive hierarchical search and evolves both the memory and retrieval policies. Each row changes one design choice. Initializing the evolution from \modelS achieves similar accuracy. Evolving only the memory policy leads to a policy that uses $2.9\times$ more tokens with similar accuracy.}
\vspace{-0.2cm}
\label{tab:abl_evolution}
\small
\setlength{\tabcolsep}{2.2pt}%
\begin{tabular}{lccgg}
\toprule
& \multicolumn{3}{c}{Input Tokens / Q} & \multicolumn{1}{c}{} \\%
\cmidrule(lr){2-4}
Method & Search & QA & Total & Avg. Acc. \\
\midrule
\modelE (default)                & 33.9K          & \textbf{4.7K} & \textbf{38.6K} & 57.4          \\
\midrule
\multicolumn{5}{l}{\textit{Initialization}} \\
\quad From \modelS                & \textbf{32.0K} & 11.7K & 43.7K & \textbf{58.1} \\
\midrule
\multicolumn{5}{l}{\textit{Evolution component}} \\
\quad Memory-only                 & 108.3K & 4.8K & 113.1K & 57.3 \\
\bottomrule
\end{tabular}
\vspace{-0.3cm}
\end{table}

\noindent\textbf{Generalization to the test benchmarks.}
In \cref{tab:policy_comparison}, we compare \modelE with the initial policy on the five test benchmarks, whose questions are never used during the evolution.
\modelE uses $5.6\times$ fewer input tokens than the initial policy ($38.6$K vs.\ $214.5$K), and most of this reduction comes from the search ($33.9$K vs.\ $207.1$K).
Its average accuracy is slightly higher ($57.4\%$ vs.\ $56.9\%$), and it is lower only on EgoMemReason.
The reduction in tokens is close to that on the validation set ($5.6\times$ vs.\ $6.5\times$), which shows that the evolved policy generalizes beyond the questions used for the evolution.

\noindent\textbf{Choice of initial policy.}
In \cref{tab:abl_evolution}, we also start the evolution from \modelS instead of the hand-designed recursive hierarchical policy.
The resulting policy is close to \modelE in both input tokens ($43.7$K vs.\ $38.6$K) and accuracy ($58.1\%$ vs.\ $57.4\%$), which shows that the evolution is robust to the initial policy.
However, \modelS uses $6.2\times$ more tokens per question than the hand-designed recursive hierarchical policy ($1.32$M vs.\ $214.5$K), which makes the early iterations costly to evaluate.
We therefore start the evolution from the recursive hierarchical policy by default.

\noindent\textbf{Joint evolution.}
We also evolve only the memory policy and keep the retrieval policy unchanged (Memory-only in \cref{tab:abl_evolution}).
This variant achieves similar accuracy to \modelE ($57.3\%$ vs.\ $57.4\%$) but uses $2.9\times$ more tokens ($113.1$K vs.\ $38.6$K), and the difference comes from the search ($108.3$K vs.\ $33.9$K).
Evolving the memory policy alone is thus not sufficient, and the two policies should be evolved jointly.

\subsection{Ablation Studies}
\label{sec:ablations}

\begin{table}[t]
\centering
\caption{\textbf{Ablation on the retrieval method} in \modelS on EgoLifeQA. Each method scores the captions of all clips and passes the top-$100$ captions to the same LLM. Hit@$K$ is the percentage of questions for which the top-$K$ retrieved clips cover the evidence. The decision model achieves the best results on all metrics.}
\vspace{-0.2cm}
\label{tab:abl_retrieval}
\small
\setlength{\tabcolsep}{3.5pt}%
\begin{tabular}{lcccc}
\toprule
Retrieval Method & Acc. & Hit@1 & Hit@10 & Hit@100 \\
\midrule
Lexical matching~\cite{robertson2009bm25} & 65.2 & 9.6 & 29.4 & 54.0 \\
Embedding similarity~\cite{zhang2025qwen3embedding} & 70.2 & 13.6 & 37.0 & 64.0 \\
Reranker~\cite{zhang2025qwen3embedding} & 71.6 & 20.2 & 48.4 & 69.8 \\
\midrule
Decision model~\cite{typesafe2026jev} & \textbf{72.6} & \textbf{24.2} & \textbf{52.6} & \textbf{75.0} \\
\bottomrule
\end{tabular}
\end{table}

\addtocounter{table}{1}
\begin{table}[t]
\centering
\caption{\textbf{Ablation on the decision model} for \modelE. We replace the default decision model, Jev~\cite{typesafe2026jev}, with Decider~\cite{perplexity2026decider}, Drex Max~\cite{dhankhar2026drex}, and OpenJev~\cite{razorback2026openjev}, and reuse the policy evolved with Jev. We report the accuracy ($\%$) averaged over the five benchmarks.}
\vspace{-0.2cm}
\label{tab:abl_decision_model}
\small
\setlength{\tabcolsep}{5pt}
\begin{tabular}{lcccc}
\toprule
Decision Model & Jev & Decider & Drex Max & OpenJev \\
\midrule
Avg. Acc. & 57.4 & 57.5 & \textbf{57.6} & 51.4 \\
\bottomrule
\end{tabular}
\end{table}

We ablate three design choices, all with GPT-6 Luna.

\noindent\textbf{Retrieval method.}
In \cref{tab:abl_retrieval}, we replace the decision model in \modelS with three standard retrieval methods, namely lexical matching (BM25~\cite{robertson2009bm25}), embedding similarity (Qwen3-Embedding-8B~\cite{zhang2025qwen3embedding}), and a reranker (Qwen3-Reranker-8B~\cite{zhang2025qwen3embedding}).
Each method scores the captions of all clips and passes the top-$100$ captions to the same LLM.
Besides the accuracy, we report Hit@$K$, the percentage of questions for which at least one of the top-$K$ retrieved clips covers the evidence.
The decision model achieves the best results on all metrics.
Compared with lexical matching and embedding similarity, it improves the accuracy by $7.4\%$ and $2.4\%$ and Hit@$100$ by $21.0\%$ and $11.0\%$.
The reranker, which also processes the question and each caption jointly, is the closest method, and the decision model still improves Hit@$100$ by $5.2\%$ and the accuracy by $1.0\%$.

\noindent\textbf{Number of retrieved captions.}
In \cref{tab:abl_num_captions}, we vary the number of retrieved captions $K$ in \modelS.
The average accuracy saturates around $K = 50$.
Increasing $K$ from $10$ to $50$ improves the average accuracy by $4.6\%$, while increasing it further to $100$ adds only $0.3\%$.
As a reference, we also report MERIT with the same LLM, which passes $106$ clips to the LLM per question on average.
With a similar number of clips, \modelS outperforms MERIT by $1.7\%$ on average, with the largest gains on EgoLifeQA ($4.0\%$) and MM-Lifelong ($7.8\%$).
On MM-Lifelong, \modelS with only $10$ captions already outperforms MERIT, which uses $154$ clips on average ($31.0\%$ vs.\ $29.1\%$).
MERIT remains higher on EgoMemReason and LVBench.
EgoMemReason requires combining evidence from $5.1$ video segments per question on average~\cite{wang2026egomemreason}, and MERIT passes $167$ clips to the LLM on this benchmark, $1.7\times$ more than \modelS.
The accuracy of \modelS on EgoMemReason also increases from $K = 50$ to $K = 100$.

\noindent\textbf{Decision model.}
In \cref{tab:abl_decision_model}, we replace Jev in \modelE with three other decision models and reuse the policy evolved with Jev without any changes.
Decider and Drex Max achieve similar accuracy to Jev, which shows that the evolved policy is not specific to the decision model that it was evolved with.
In contrast, OpenJev is $6.0\%$ lower, which indicates that the accuracy of \modelE depends on the capability of the decision model.

\section{Conclusion}
\label{sec:conclusion}
We presented \model, a framework that leverages pretrained decision models to retrieve evidence from ultra-long videos.
Our first method, \modelS, scores every clip caption with the decision model and passes the top-ranked captions to an LLM.
Despite its simplicity, \modelS achieves state-of-the-art accuracy on three ultra-long video understanding benchmarks, EgoLifeQA, EgoMemReason, and MM-Lifelong, and generalizes to the long-video benchmarks VideoMME-L and LVBench.
To reduce the cost of \modelS, we further proposed an automated policy evolution framework, in which a coding agent jointly revises the memory and retrieval policies around a fixed decision model.
The resulting method, \modelE, maintains comparable accuracy while reducing input tokens by $34\times$ on average compared with \modelS.
It is also $3.8\times$ faster than the strongest prior method on EgoLifeQA while achieving higher accuracy.
Together, our results suggest that pretrained decision models are an accurate and efficient alternative to similarity-based retrieval and LLM-based search for ultra-long video understanding.
We hope that our work encourages future research on decision models for video understanding.

\iftoggle{cvprfinal}{\section*{Acknowledgements}
\label{sec:acknowledgements}
This work was supported by the Laboratory for Analytic Sciences via NC State University, ONR Award N00014-23-1-2356, the Sony Focused Research Award, NSF CAREER Award 2541848, and the Home Team Science and Technology Agency (HTX), Singapore, through a research collaboration with ST Engineering.
}{}

{
    \small
    \bibliographystyle{ieeenat_fullname}
    \bibliography{main}
}

\makeatletter
\immediate\write\@auxout{\string\newlabel{main:lastfigure}{{\arabic{figure}}{}{}{}{}}}
\immediate\write\@auxout{\string\newlabel{main:lasttable}{{\arabic{table}}{}{}{}{}}}
\makeatother
\fi

\clearpage
\setcounter{page}{1}
\maketitlesupplementary

\appendix
\crefname{appendix}{Sec.}{Secs.}
\Crefname{appendix}{Section}{Sections}
\crefname{subappendix}{Sec.}{Secs.}
\Crefname{subappendix}{Section}{Sections}

\setcounter{dbltopnumber}{4}
\makeatletter
\setlength{\@dblfptop}{0pt}
\setlength{\@fptop}{0pt}
\makeatother

\providecommand{\ptus}{\char95 }
\providecommand{\ptbt}{\char96 }
\providecommand{\ptlb}{\char123 }
\providecommand{\ptrb}{\char125 }
\providecommand{\promptbox}[1]{%
  \fcolorbox{black!30}{black!3}{%
    \begin{minipage}{\dimexpr\linewidth-2\fboxsep-2\fboxrule\relax}%
      \fontfamily{cmtt}\fontsize{7}{8.3}\selectfont\raggedright
      \setlength{\parskip}{3pt}\setlength{\parindent}{0pt}#1%
    \end{minipage}}}
\providecommand{\figplaceholder}[2]{%
  \fcolorbox{black!40}{black!4}{%
    \begin{minipage}[c][#1][c]{\dimexpr\linewidth-2\fboxsep-2\fboxrule\relax}%
      \centering\small\color{red}#2%
    \end{minipage}}}

\noindent This supplementary material is organized as follows.
We first provide additional implementation details (\cref{sec:supp_impl}) and introduce the evaluation benchmarks (\cref{sec:supp_benchmarks}).
We then describe the automated policy evolution, including the initial policy and the development and validation sets (\cref{sec:supp_evolution}), and the evolved policy \modelE (\cref{sec:supp_delovi_e}).
Lastly, we present additional results (\cref{sec:supp_analysis}) and qualitative results (\cref{sec:supp_qual}).

\section{Additional Implementation Details}
\label{sec:supp_impl}

\noindent\textbf{Captioning and speech transcription.}
\label{sec:supp_captioning}
For EgoLifeQA and EgoMemReason, we use the official captions and transcripts of EgoLife~\cite{yang2025egolife}, which describe clips of $30$ seconds.
For the other benchmarks, we divide each video into clips of $10$ seconds (VideoMME-L and LVBench) or $30$ seconds (MM-Lifelong).
For each clip, we sample one frame per second and prompt GPT-6 Luna~\cite{openai2026gpt6luna} to describe the visual content of the clip, with low image detail and without reasoning.
We transcribe the speech of each video with Whisper~\cite{radford2023whisper}.
The transcript lines that overlap with a clip are appended to its caption.
The caption, the transcript, and the start and end time of the clip form one memory entry.

\noindent\textbf{Decision model.}
\label{sec:supp_models}
We use Jev~\cite{typesafe2026jev} (version $1.13.0$) as the decision model $\mathcal{D}$ and use two types of requests.
In a \emph{score} request, $\mathcal{D}$ receives the question and a single candidate and outputs a relevance score in $[0, 1]$.
\modelS uses score requests to score every memory entry (\cref{eq:delovi_s}).
In a \emph{ranking} request, $\mathcal{D}$ receives the question and a set of candidates, processes all candidates jointly, and outputs a selection probability for each candidate, which we use as its relevance score.
The initial policy and \modelE use ranking requests to rank the nodes of the memory tree.

\noindent\textbf{LLMs.}
By default, the LLM is GPT-6 Luna with its default reasoning effort.
GPT-6 Luna also writes all summaries in the memory of the initial policy and of \modelE.
In \cref{tab:main}, we also report results with GPT-5~\cite{openai2025gpt5} and GPT-6.1 Sol~\cite{openai2026gpt61sol}.

\noindent\textbf{Prompts.}
\label{sec:supp_prompts}
We provide the prompts for memory construction in \cref{fig:supp_prompt_memory}, which include the prompt of the captioner and the prompts of the summaries in the memory.
In \cref{fig:supp_prompt_requests}, we provide the score request of \modelS and the ranking request of the initial policy and \modelE.

\begin{figure*}[t]
\centering
\begin{minipage}[t]{0.49\linewidth}
{\footnotesize\textbf{(a) Captioner}}\\[2pt]
\promptbox{%
You are an expert video captioner.\par
You will receive a short video segment represented by ordered frames.\\
Write a caption describing the visual content of the segment.\par
Guidelines:\\
- Describe visible actions, people, objects, and environment.\\
- Keep the caption factual and neutral.\\
- Do not mention frames, timestamps, or that the input came from frames.\\
- Avoid speculation about emotions or intentions unless clearly visible.\par
Output only the final caption text.}\\[6pt]
{\footnotesize\textbf{(c) Timeline of an hour: one line per 3-minute window}}\\[2pt]
\promptbox{%
You write one line of a timeline of a recording: what happened in one segment of about three minutes. You are given the segment's caption or summary. The line is read to find when something happened, in what order, how often, and what else happened at the same time.\par
Say who did what, by name, in time order, separated by semicolons; when several people act at once, give each person's action ("Jake washes the pot while Tasha and Lucia discuss the cake"). Keep the place, distinctive objects, numbers and counts ("opens the oven twice", "pays 50 yuan"), on-screen text, games and titles, and the topic or outcome of what was said. Keep the point of view of the text: if it is written in the first person ("I"), keep "I"; otherwise refer to people by name or role. Facts only, nothing the text does not say. No introduction, no time stamps.\par
At most \ptlb max\ptus words\ptrb{} words.}\\[6pt]
{\footnotesize\textbf{(d) Index of a day: one line per hour}}\\[2pt]
\promptbox{%
You write one line of an index of a long recording. You are given the caption or summary of one segment. The line is read to decide whether a specific moment that a question asks about lies in this segment.\par
List the segment's events in time order, separated by semicolons, each with its specific, searchable details: the people involved, by name; the place; distinctive objects; the activity; games, titles, on-screen text and numbers; the topics of what was said. Keep the point of view of the text: if it is written in the first person ("I"), keep "I"; otherwise refer to people by name or role. Facts only, nothing the text does not say. No full sentences, no introduction, no time stamps.\par
At most \ptlb max\ptus words\ptrb{} words.}
\end{minipage}\hfill
\begin{minipage}[t]{0.49\linewidth}
{\footnotesize\textbf{(b) Free-text summary}}\\[2pt]
\promptbox{%
You summarize the captions of consecutive segments of one recording into an event summary. The summary is used to decide whether a specific moment lies inside this span of the recording, and as context for answering questions about it.\par
\#\#\#\# Requirements\par
1. **Structure the events clearly**\\
- Merge related content into major events and arrange them in chronological order.\\
- Group events by location, task, or theme. Each event has a clear start, progression, and key turning points.\par
2. **Retain key information**\\
- The actions and decisions of the people involved, by name. Keep the captions' point of view: if they are written in the first person ("I"), "I" is the camera wearer, so keep "I"; otherwise refer to people by name or role (for example "the streamer").\\
- Discussions, decisions, plans, and task arrangements that matter, with their outcome.\\
- The purpose and method of key actions (for example "orders takeout on a phone", "writes the plan on a whiteboard").\\
- The specific, searchable details: names, distinctive objects, places, on-screen text, titles, numbers and prices, and the topics of what was said.\par
3. **Be concise**\\
- Keep the facts clear; leave out atmosphere, emotions, and abstract commentary.\\
- Drop trivial chatter; keep the core topic and conclusion of each discussion.\par
4. **Strictly adhere to the facts**\\
- Do not assume or add anything the captions do not say.\\
- Keep the true chronological order of events.\par
\#\#\#\# Output format\\
One paragraph per major event: start with one sentence naming the event, then the key actions and details, then its outcome. At most \ptlb max\ptus words\ptrb{} words in total. Do not mention the date, the time, or the word count.}
\end{minipage}
\caption{\textbf{Prompts for memory construction.} \textbf{(a)} Prompt of the captioner. \textbf{(b)} Prompt of the free-text summaries of the initial policy and of \modelE. \textbf{(c)} One line of the timeline of an hour in \modelE, written from the summary of one 3-minute window. \textbf{(d)} One line of the index of a day in \modelE, written from the summary of one hour.}
\label{fig:supp_prompt_memory}
\end{figure*}

\begin{figure*}[t]
\centering
\begin{minipage}[t]{0.49\linewidth}
{\footnotesize\textbf{(a) Score request of \modelS}}\\[2pt]
\promptbox{%
\textrm{\textit{Input fields:}}\\
question: <question>\\
answer\ptus options: <answer options, if any>\\
memory: <text of one memory entry>\par
\textrm{\textit{Instruction:}}\\
\ptbt memory\ptbt{} describes a 30-second clip of first-person video from Jake's head-mounted camera. "I" in \ptbt memory\ptbt{} and in \ptbt question\ptbt{} both refer to Jake. Jake asked \ptbt question\ptbt{} at a later time. \ptbt answer\ptus options\ptbt{} lists its candidate answers. How useful is \ptbt memory\ptbt{} as evidence for answering \ptbt question\ptbt?\par
\textrm{\textit{Levels:}}\\
0: \ptbt memory\ptbt{} involves different people, objects, places, and activities from the ones \ptbt question\ptbt{} asks about.\\
1: \ptbt memory\ptbt{} shares the setting, people, or general activity of \ptbt question\ptbt, but does not mention the specific object, action, or event that \ptbt question\ptbt{} asks about.\\
2: \ptbt memory\ptbt{} mentions the specific object, action, or event that \ptbt question\ptbt{} asks about, but not the detail needed to answer \ptbt question\ptbt.\\
3: \ptbt memory\ptbt{} shows the moment that \ptbt question\ptbt{} asks about, with the detail needed to answer \ptbt question\ptbt.}
\end{minipage}\hfill
\begin{minipage}[t]{0.49\linewidth}
{\footnotesize\textbf{(b) Ranking request of the initial policy and \modelE}}\\[2pt]
\promptbox{%
\textrm{\textit{Input fields:}}\\
question: <question>\\
answer\ptus options: <answer options, if any>\\
question\ptus time: <time at which the question is asked, if any>\\
memory: <description of the video>\par
\textrm{\textit{Instruction:}}\\
\ptbt question\ptbt{} asks about the recording described in \ptbt memory\ptbt. \ptbt answer\ptus options\ptbt{} lists its candidate answers. The question was asked at \ptbt question\ptus time\ptbt; every segment below ended before then. Each choice is the caption of one segment of the recording. Segments differ in length; the caption of a long segment is a summary, so the moment \ptbt question\ptbt{} asks about can lie inside a long segment even when its caption mentions it only briefly. Which segment contains the moment that answers \ptbt question\ptbt?\par
\textrm{\textit{Each candidate:}}\\
time: <start and end time of the node>\\
duration: <duration of the node>\\
caption: <text of the node>}
\end{minipage}
\caption{\textbf{Requests to the decision model.} \textbf{(a)} In a score request, the decision model receives the question and one memory entry and outputs a probability for each of the four levels. The relevance score is the expected level, normalized to $[0, 1]$. We show the instruction for EgoLifeQA, and its description of the video is adapted to each benchmark. \textbf{(b)} In a ranking request, the decision model receives the input fields, the instruction, and the candidates, and outputs a selection probability for each candidate.}
\label{fig:supp_prompt_requests}
\end{figure*}

\section{Evaluation Benchmarks}
\label{sec:supp_benchmarks}

\noindent\textbf{EgoLifeQA}~\cite{yang2025egolife} contains multiple-choice questions on the week-long egocentric videos of EgoLife, in which six participants live together for one week.
Following prior works~\cite{yeo2026worldmm, choi2026merit}, we evaluate on the $500$ questions on the video of one participant.
Each question is asked at a specific time of the week and has to be answered from the video recorded before this time, which ranges from $0.2$ to $51.9$ hours.

\noindent\textbf{EgoMemReason}~\cite{wang2026egomemreason} contains $500$ multiple-choice questions on the videos of all six EgoLife participants.
Its questions require reasoning over long-term memory, \eg, ordering events, counting, and tracking the state of objects, and combine evidence from $5.1$ video segments per question on average.

\noindent\textbf{MM-Lifelong}~\cite{chen2026mmlifelong} contains $181.1$ hours of video at the day, week, and month scales.
Its questions are open-ended.
Following the official protocol, we evaluate on the test splits of the day and week scales ($200$ questions each) and on the validation split of the month scale ($623$ questions).
We use the official rubric-based grader with GPT-5 as the judge and report its score averaged over all $1{,}023$ questions.

\noindent\textbf{VideoMME-L} is the long split of Video-MME~\cite{fu2025videomme}, which contains $300$ videos of $30$ to $60$ minutes with $900$ multiple-choice questions.

\noindent\textbf{LVBench}~\cite{wang2025lvbench} contains $1{,}549$ multiple-choice questions on hour-long videos.

\section{Automated Policy Evolution}
\label{sec:supp_evolution}

We describe the initial policy (\cref{sec:supp_hier_search}, \cref{alg:search}), the setup of the evolution (\cref{sec:supp_evolution_setup}), and the development and validation sets (\cref{sec:supp_devval}) of the automated policy evolution (\cref{sec:delovi_e} of the main paper).

\begin{algorithm}[b]
\caption{Recursive hierarchical search}
\label{alg:search}
\small
\renewcommand{\algorithmicrequire}{\textbf{Input:}}
\renewcommand{\algorithmicensure}{\textbf{Output:}}
\algtext*{EndFor}
\algtext*{EndIf}
\begin{algorithmic}[1]
\Require question $Q$, memory tree $\mathcal{M}_t$, decision model $\mathcal{D}$, LLM $G$
\Ensure answer $A$
\Statex \textbf{Notation:} $\textsc{Rank}_{\mathcal{D}}(Q, \mathcal{N})$ scores every node in $\mathcal{N}$ with $\mathcal{D}$ and returns the nodes sorted by relevance score, highest first.
\State $\mathcal{E} \gets \emptyset$ \Comment{evidence set}
\State $\mathcal{L} \gets \{\textsc{Rank}_{\mathcal{D}}(Q, \text{top-level nodes of } \mathcal{M}_t)\}$ \Comment{ranked lists}
\For{$i = 1$ \textbf{to} $50$}
    \State $\mathcal{C} \gets$ top-ranked node of every non-empty ranked list in $\mathcal{L}$
    \State $n \gets$ first node of $\textsc{Rank}_{\mathcal{D}}(Q, \mathcal{C})$ \Comment{most relevant}
    \State Remove $n$ from its ranked list
    \State $\mathcal{E} \gets \mathcal{E} \cup \{n\}$ \Comment{add $n$ to the evidence set}
    \If{$i \bmod 5 = 0$} \Comment{check every $5$ steps}
        \State $A, c \gets G(Q, \mathcal{E})$ \Comment{answer and confidence}
        \If{$c > 0.8$} \textbf{break}
        \EndIf
    \EndIf
    \If{$n$ has children}
        \State $\mathcal{L} \gets \mathcal{L} \cup \{\textsc{Rank}_{\mathcal{D}}(Q, \text{children of } n)\}$
    \EndIf
\EndFor
\State \Return $A$
\end{algorithmic}
\end{algorithm}

\subsection{Initial Policy: Recursive Hierarchical Search}
\label{sec:supp_hier_search}
The evolution starts from a hand-designed policy, which we refer to as recursive hierarchical search.

\noindent\textbf{Memory policy.}
We divide the video into short clips as it is observed and describe each clip with a visual caption and a speech transcript, as in \modelS.
Whenever a 3-minute window of the video is complete, an LLM summarizes the clips in this window.
The 3-minute summaries are then summarized into 15-minute summaries, and the 15-minute summaries into 1-hour summaries.
The memory is therefore a tree, where the leaves are the clip-level memory entries and the internal nodes are the summaries.

\noindent\textbf{Retrieval policy.}
We use the decision model $\mathcal{D}$ to search this tree (\cref{alg:search}).
At a high level, the search starts from the top-level nodes and recursively expands the most relevant parts of the tree.
At each step, it selects the node that is most relevant to the question, adds it to the evidence set $\mathcal{E}$, and makes its children available for selection.
Therefore, at each step the search can either go deeper into a selected node or move to a different part of the video.
Specifically, given the question $Q$ and a set of nodes $\mathcal{N}$, $\mathcal{D}$ outputs a relevance score for each node, which indicates how likely the node is to contain the moment that answers the question.
We denote by $\textsc{Rank}_{\mathcal{D}}(Q, \mathcal{N})$ the nodes of $\mathcal{N}$ sorted by their relevance scores, from the highest to the lowest.
The search maintains a set of \emph{ranked lists} $\mathcal{L}$, where each ranked list contains the children of one node sorted by their relevance scores.
We first rank the top-level nodes of the tree with $\mathcal{D}$, which gives the first ranked list.
At each step, we collect the top-ranked node of every ranked list as the candidates $\mathcal{C}$.
$\mathcal{D}$ then ranks the candidates, and we select the one with the highest relevance score.
The selected node is removed from its ranked list and added to the evidence set.
If it has children, we rank them with $\mathcal{D}$ to form a new ranked list.
Every $5$ steps, the LLM $G$ answers the question from the evidence set and reports its confidence.
The search stops once the confidence exceeds $0.8$ or after $50$ steps, and the last answer is the final prediction.

\subsection{Evolution Setup}
\label{sec:supp_evolution_setup}

\noindent\textbf{Policies.}
A policy consists of two executable programs.
The memory policy receives the memory entries of a video in temporal order and builds the memory.
The retrieval policy receives the question and the memory, sends requests to the decision model $\mathcal{D}$, calls the LLM, and returns the answer.
The coding agent can change everything inside these two programs, while the captioner, the decision model $\mathcal{D}$, the LLM with its prompt, and the evaluation are fixed.
Every policy has to build its memory online.
Specifically, a node of the memory is only created after all of its clips have been observed, existing nodes are never changed, and a question asked at time $t$ only has access to the nodes that existed at $t$.

\noindent\textbf{Objective.}
We evaluate each policy on the development set by its accuracy and its cost, \ie, the number of input tokens per question.
When comparing the cost of two policies during the evolution, we give the input tokens of the LLM a higher weight than those of the decision model, since they are more expensive.
We also discard every policy that is notably less accurate than the initial policy, regardless of its cost.

\begin{table*}[t]
\centering
\caption{\textbf{Development and validation sets} of the automated policy evolution. For each dataset, we list the video duration, the question types, and the number of questions in the two sets. All questions are multiple-choice, except for the open-ended questions of MM-Lifelong. For MM-Lifelong, we use the official dev split.}
\label{tab:supp_devval}
\small
\setlength{\tabcolsep}{5pt}
\begin{tabular}{ll>{\raggedright\arraybackslash}p{8.6cm}cc}
\toprule
Dataset & Video Duration & Question Types & Dev. & Val. \\
\midrule
Ego-R1~\cite{tian2025egor1} & one week & entity log, event recall, habit insight, relation map, task master & 75 & 50 \\
MA-EgoQA~\cite{kim2026maegoqa} & one week & social interaction, task coordination, theory of mind, temporal reasoning, environmental interaction & 120 & 80 \\
MM-Lifelong~\cite{chen2026mmlifelong} (dev split) & one month & 11 types, \eg, counting, causal reasoning, and entity recognition & 75 & 50 \\
LongVideoBench~\cite{wu2024longvideobench} & 30 to 60 min & referring reasoning (perception and relation) & 50 & 30 \\
HourVideo~\cite{chandrasegaran2024hourvideo} & 30 to 120 min & perception, visual reasoning, summarization & 70 & 50 \\
VRBench~\cite{yu2025vrbench} & 60 to 180 min & event attribution, event summarization, counting, implicit inference, logical linkage, event prediction & 60 & 40 \\
HD-EPIC~\cite{perrett2025hdepic} & 3.5 to 7 hours & recipe, ingredient, nutrition & 50 & 50 \\
\midrule
Total & & & 500 & 350 \\
\bottomrule
\end{tabular}
\end{table*}

\noindent\textbf{Coding agent.}
We use Claude Code~\cite{anthropic2025claudecode} with Claude Opus 5.5~\cite{anthropic2026opus55} at high reasoning effort.
In each iteration, the agent starts a new session with the programs of the parent policy.
Its instructions state the objective, the results of the parent policy and of all earlier policies, and a set of rules.
The agent can inspect the programs and the development results of every earlier policy, including the retrieval trajectory of every development question.
A retrieval trajectory records every request to $\mathcal{D}$ with its candidates and relevance scores, every answer of the LLM with its confidence, and the evidence set.
The rules instruct the agent not to fit the policy to the development questions and not to make dataset-specific changes.
In particular, a policy may only base its decisions on properties of the video and of the question, \eg, the duration of the video, and never on the name of a dataset.
After the agent finishes, the new policy is evaluated on the whole development set.

\noindent\textbf{Parent selection.}
After each iteration, a second call of the same agent selects the parent policy of the next iteration from the Pareto frontier.
It is instructed to select the policy whose revision is most likely to extend the Pareto frontier.

\noindent\textbf{Validation set and final policy.}
The coding agent never has access to the validation questions.
We evaluate a policy on the validation set if it is on the Pareto frontier of the development set when it is proposed.
After the last iteration, we select the policy on the Pareto frontier with the highest validation accuracy as \modelE.
We run the evolution for $30$ iterations.
In the later iterations, the cost converges and the accuracy no longer improves (\cref{fig:evolution}).

\subsection{Development and Validation Sets}
\label{sec:supp_devval}

We list the development and validation sets in \cref{tab:supp_devval}.
They contain $500$ and $350$ questions from seven datasets: Ego-R1~\cite{tian2025egor1}, MA-EgoQA~\cite{kim2026maegoqa}, MM-Lifelong~\cite{chen2026mmlifelong}, LongVideoBench~\cite{wu2024longvideobench}, HourVideo~\cite{chandrasegaran2024hourvideo}, VRBench~\cite{yu2025vrbench}, and HD-EPIC~\cite{perrett2025hdepic}.
We select these datasets to cover different video durations and domains, from single videos of $30$ minutes to videos that span a week or a month, so that the evolved policy is not specific to one type of video.
For MM-Lifelong, we use the official dev split.
For each dataset, we also choose questions of diverse types (\cref{tab:supp_devval}), which range from recalling a single entity or event to counting, temporal and causal reasoning, and summarization.
We split the questions of each dataset into the two sets at random.
For the multiple-choice questions, we report the accuracy.
For the open-ended questions of MM-Lifelong, an LLM judge compares each answer with the reference answer.
The questions of both sets do not overlap with the questions of our evaluation benchmarks.

\section{\modelE: The Evolved Policy}
\label{sec:supp_delovi_e}
In this section, we provide the details about the memory tree and the retrieval policy of \modelE.
Specifically, we describe the format of the summaries at each level of the tree, the full algorithm (\cref{alg:retrieval}), the details of the hierarchical search and the clip re-ranking, and the complete stopping rule.

\begin{algorithm}[t]
\caption{\modelE retrieval and question answering}
\label{alg:retrieval}
\small
\renewcommand{\algorithmicrequire}{\textbf{Input:}}
\renewcommand{\algorithmicensure}{\textbf{Output:}}
\algtext*{EndFor}
\algtext*{EndIf}
\begin{algorithmic}[1]
\Require question $Q$, video $V_{<t}$, memory tree $\mathcal{M}_t$, decision model $\mathcal{D}$, LLM $G$
\Ensure answer $A$
\Statex \textbf{Notation:} $\textsc{Rank}_{\mathcal{D}}(Q, \mathcal{N})$ returns the nodes in $\mathcal{N}$ sorted by relevance score, as in \cref{alg:search}.
\State $\mathcal{E} \gets \emptyset$ \Comment{evidence set}
\Statex \textit{// Step 1: find the first evidence window by hierarchical search}
\State $\mathcal{L} \gets \{\textsc{Rank}_{\mathcal{D}}(Q, \text{top-level nodes of } \mathcal{M}_t)\}$ \Comment{ranked lists}
\Repeat
    \State $\mathcal{C} \gets$ top-ranked node of every non-empty ranked list in $\mathcal{L}$
    \State $n \gets$ first node of $\textsc{Rank}_{\mathcal{D}}(Q, \mathcal{C})$ \Comment{most relevant}
    \State Remove $n$ from its ranked list
    \If{$n$ is not a 3-minute window}
        \State $\mathcal{L} \gets \mathcal{L} \cup \{\textsc{Rank}_{\mathcal{D}}(Q, \text{children of } n)\}$
    \EndIf
\Until{$n$ is a 3-minute window}
\State $\mathcal{E} \gets \mathcal{E} \cup \{$short clips of $n$, summary of its hour$\}$
\State $A, c \gets G(Q, \mathcal{E})$ \Comment{answer and confidence}
\If{$c > 0.8$} \Return $A$ \Comment{stop if confident}
\EndIf
\Statex \textit{// Step 2: re-rank candidate short clips with $\mathcal{D}$}
\State $\mathcal{C} \gets \textsc{BM25}(Q, \mathcal{M}_t)$ \Comment{short clips matching $Q$}
\State sort $\mathcal{C}$ with $\mathcal{D}$ \Comment{by relevance to $Q$}
\Statex \textit{// Step 3: add evidence windows until the answer is confident}
\State $W \gets \textsc{MaxWindows}(Q, V_{<t})$ \Comment{max.\ evidence windows}
\For{$k = 2$ \textbf{to} $W$}
    \State $m \gets$ next short clip in $\mathcal{C}$
    \State $\mathcal{E} \gets \mathcal{E} \cup \textsc{Window}(m)$ \Comment{$m$ and its neighbors}
    \If{$k = 2^n$ \textbf{or} $k = W$}
        \State $A, c \gets G(Q, \mathcal{E})$ \Comment{answer and confidence}
        \If{$c > 0.8$} \Return $A$
        \EndIf
    \EndIf
\EndFor
\State \Return $A$ \Comment{answer after $W$ windows}
\end{algorithmic}
\end{algorithm}

\begin{table*}[t]
\centering
\caption{\textbf{Per-benchmark results of the evolution ablations.} The default \modelE is initialized from the recursive hierarchical search and evolves both the memory and retrieval policies. Each of the other two rows changes one design choice. All policies use the same LLM (GPT-6 Luna). We report the accuracy ($\%$) on each benchmark and the input tokens per question, averaged over the five benchmarks. Both variants achieve similar accuracy to \modelE, and evolving only the memory policy leads to a policy that uses $2.9\times$ more tokens.}
\label{tab:supp_abl_evolution}
\small
\setlength{\tabcolsep}{3pt}
\begin{tabular}{lccgcccccg}
\toprule
& \multicolumn{3}{c}{Input Tokens / Q} & \multicolumn{6}{c}{Accuracy} \\
\cmidrule(lr){2-4} \cmidrule(lr){5-10}
Method & Search & QA & Total & EgoLifeQA & EgoMemReason & MM-Lifelong & VideoMME-L & LVBench & Avg. \\
\midrule
\modelE (default) & 33.9K & \textbf{4.7K} & \textbf{38.6K} & 69.6 & 37.6 & 34.5 & 77.2 & 68.1 & 57.4 \\
\midrule
\multicolumn{10}{l}{\textit{Initialization}} \\
\quad From \modelS & \textbf{32.0K} & 11.7K & 43.7K & \textbf{72.0} & 36.8 & \textbf{35.4} & 77.6 & \textbf{68.5} & \textbf{58.1} \\
\midrule
\multicolumn{10}{l}{\textit{Evolution component}} \\
\quad Memory-only & 108.3K & 4.8K & 113.1K & 66.8 & \textbf{39.8} & 33.9 & \textbf{77.8} & 68.4 & 57.3 \\
\bottomrule
\end{tabular}
\end{table*}

\begin{table*}[t]
\centering
\caption{\textbf{Per-benchmark results of the decision model ablation} for \modelE. We replace the default decision model, Jev~\cite{typesafe2026jev}, with Decider~\cite{perplexity2026decider}, Drex Max~\cite{dhankhar2026drex}, and OpenJev~\cite{razorback2026openjev}, and reuse the policy evolved with Jev without any changes. All rows use the same LLM (GPT-6 Luna). We report the accuracy ($\%$). Decider and Drex Max are within $2.3\%$ of Jev on every benchmark.}
\label{tab:supp_abl_decision_model}
\small
\setlength{\tabcolsep}{7pt}
\begin{tabular}{lcccccg}
\toprule
Decision Model & EgoLifeQA & EgoMemReason & MM-Lifelong & VideoMME-L & LVBench & Avg. \\
\midrule
Jev (default) & \textbf{69.6} & 37.6 & 34.5 & 77.2 & 68.1 & 57.4 \\
Decider & 69.0 & 37.2 & 35.2 & \textbf{77.3} & 68.7 & 57.5 \\
Drex Max & \textbf{69.6} & \textbf{39.0} & \textbf{35.3} & 74.9 & \textbf{69.2} & \textbf{57.6} \\
OpenJev & 52.8 & 34.2 & 32.4 & 74.8 & 62.6 & 51.4 \\
\bottomrule
\end{tabular}
\end{table*}

\noindent\textbf{Memory policy.}
The summaries have a different format at each level of the tree.
The summary of a 3-minute window describes the events in its clips in free text.
The summary of an hour is a timeline with one timestamped line per 3-minute window, since a timeline preserves the order and frequency of events better than free text.
The summary of a day is an index with one line per hour, which allows the decision model $\mathcal{D}$ to first locate the relevant day instead of reading every hour of the video.
For videos that span multiple days, the memory also stores a free-text summary of each hour.
The decision model $\mathcal{D}$ reads this summary to judge whether a moment lies in the hour, while the LLM reads the timeline to answer the question.
For a single video of up to two hours, one timeline covers the whole video, and the LLM reads it for every question.

We summarize the retrieval and question answering of \modelE in \cref{alg:retrieval}.

\noindent\textbf{Hierarchical search.}
The hierarchical search (Step~1 of \cref{alg:retrieval}) maintains a set of ranked lists, as in the recursive hierarchical search (\cref{alg:search}).
Specifically, $\mathcal{D}$ ranks the children of every selected node into a ranked list, and the top-ranked nodes of all ranked lists are compared at each step.
Therefore, the search can also move to a different part of the tree instead of only going deeper.
Once the selected node is a 3-minute window, all of its clips are added to the evidence set $\mathcal{E}$ together.

\noindent\textbf{Clip re-ranking.}
BM25 matches the clip captions against the question and its answer options, and $\mathcal{D}$ re-ranks the top-scoring clips in a single request.
Each of these clips forms an evidence window together with the clips within $90$ seconds before and after it, which are added from the nearest clip on until the window reaches $4$K characters.
The clip re-ranking complements the hierarchical search, which reaches a clip only through the summaries above it and can therefore miss the relevant moment when it selects the wrong day or hour.
The clip re-ranking only runs when the answer from the first evidence window is not confident, which saves the cost of ranking the BM25 clips with $\mathcal{D}$ for many questions.

\noindent\textbf{Stopping rule and the values of $W$.}
We set the confidence threshold to $0.8$.
The maximum number of evidence windows $W$ is $8$ by default.
In this case, the LLM answers after the 1st, 2nd, and 4th evidence windows, and gives the final answer after the 8th evidence window.
The checks become sparser as the search proceeds, since each check reads all the evidence again.
Besides the confidence threshold, the search also stops when the answer is the same as in the previous check.
This rule only applies from the 4th evidence window on, since the first two evidence windows often cover the same part of the video.
An answer stating that the evidence does not contain the required information never stops the search.
Questions that require several moments (\eg, a count or the last occurrence of an event) are detected from their wording and are not answered from the first evidence window alone, since a single occurrence often leads to a confident but wrong answer.
For these questions, the third evidence window also comes from the hierarchical search, and the search stops on confidence only from the 4th evidence window on.

\begin{table*}[t]
\centering
\caption{\textbf{Results on each question category of EgoLifeQA.} We report the accuracy ($\%$), with the number of questions in parentheses. The methods are grouped by their LLM, and the first group contains the methods in \cref{tab:efficiency}. Best results with each LLM are in \textbf{bold}.}
\label{tab:supp_egolifeqa_category}
\small
\setlength{\tabcolsep}{7pt}
\begin{tabular}{lcccccg}
\toprule
Method & EntityLog & EventRecall & HabitInsight & RelationMap & TaskMaster & All \\
& (125) & (126) & (61) & (125) & (63) & (500) \\
\midrule
\multicolumn{7}{l}{\textit{GPT-6 Luna}} \\
EM$^2$Mem~\cite{chen2026em2mem} & 54.4 & 66.7 & 52.5 & 61.6 & 58.7 & 59.6 \\
WorldMM~\cite{yeo2026worldmm} & 60.0 & 71.4 & 67.2 & 63.2 & 65.1 & 65.2 \\
SiLVR~\cite{zhang2026silvr} & 56.8 & 61.1 & 75.4 & 67.2 & 76.2 & 65.2 \\
MERIT~\cite{choi2026merit} & 67.8 & 68.0 & 72.8 & 63.8 & \textbf{76.8} & 68.6 \\
\oursrowS
\modelS (ours) & 67.2 & \textbf{76.2} & \textbf{80.3} & \textbf{71.2} & 71.4 & \textbf{72.6} \\
\oursrowE
\modelE (ours) & \textbf{69.6} & 69.0 & 72.1 & 68.8 & 69.8 & 69.6 \\
\midrule
\multicolumn{7}{l}{\textit{GPT-6.1 Sol}} \\
\oursrowS
\modelS (ours) & \textbf{79.2} & \textbf{75.4} & \textbf{78.7} & \textbf{77.6} & \textbf{77.8} & \textbf{77.6} \\
\oursrowE
\modelE (ours) & 70.4 & 71.4 & 72.1 & 76.8 & \textbf{77.8} & 73.4 \\
\bottomrule
\end{tabular}
\end{table*}

\section{Additional Results}
\label{sec:supp_analysis}

\noindent\textbf{Evolution ablations.}
\label{sec:supp_evo_ablation}
In \cref{tab:supp_abl_evolution}, we report the per-benchmark results of the two ablations in \cref{tab:abl_evolution} of the main paper.
The policy evolved from \modelS is within $2.4\%$ of \modelE on every benchmark, which confirms that the evolution is robust to the initial policy.
The policy obtained by only evolving the memory policy is within $2.8\%$ of \modelE on every benchmark, but its search uses $3.2\times$ more input tokens, since its retrieval policy is still the recursive hierarchical search.

\noindent\textbf{Decision model.}
\label{sec:supp_other_ablations}
In \cref{tab:supp_abl_decision_model}, we report the per-benchmark results of the decision model ablation in \cref{tab:abl_decision_model} of the main paper.
Decider and Drex Max are within $2.3\%$ of Jev on every benchmark.
With OpenJev, most of the difference comes from EgoLifeQA ($52.8\%$ vs.\ $69.6\%$).

\noindent\textbf{Results by question category on EgoLifeQA.}
\label{sec:supp_per_category}
In \cref{tab:supp_egolifeqa_category}, we report the accuracy on the five question categories of EgoLifeQA, which ask about objects (EntityLog), past events (EventRecall), personal habits (HabitInsight), the interactions among people (RelationMap), and plans and tasks (TaskMaster).
We compare \model with the four baselines of \cref{tab:efficiency}, which all use GPT-6 Luna, and also report \model with GPT-6.1 Sol.
First, with GPT-6 Luna, \modelS achieves the best accuracy on EventRecall, HabitInsight, and RelationMap, where it outperforms the best baseline by $4.0\%$ to $4.9\%$.
Second, \modelE achieves the best accuracy on EntityLog.
TaskMaster is the only category on which a baseline outperforms both \modelS and \modelE.
Lastly, with GPT-6.1 Sol, \modelE is within $0.8\%$ of \modelS on RelationMap and TaskMaster, and the largest difference between the two is on EntityLog ($70.4\%$ vs.\ $79.2\%$).

\begin{table}[t]
\centering
\caption{\textbf{Results on each split of MM-Lifelong.} We report the score ($\%$) of the official grader, with the number of questions in parentheses. The methods are grouped by their LLM. Overall is averaged over all $1{,}023$ questions, as in \cref{tab:main}. Best results with each LLM are in \textbf{bold}.}
\label{tab:supp_mml_split}
\small
\setlength{\tabcolsep}{6pt}
\begin{tabular}{lcccg}
\toprule
Method & Day & Week & Month & Overall \\
& (200) & (200) & (623) & (1,023) \\
\midrule
\multicolumn{5}{l}{\textit{GPT-5}} \\
WorldMM~\cite{yeo2026worldmm} & 12.1 & 16.2 & 13.6 & 13.8 \\
MERIT~\cite{choi2026merit} & 16.0 & 37.2 & 30.8 & 29.2 \\
\oursrowS
\modelS (ours) & \textbf{16.3} & \textbf{44.9} & \textbf{41.3} & \textbf{37.1} \\
\oursrowE
\modelE (ours) & 15.6 & 41.9 & 36.8 & 33.7 \\
\midrule
\multicolumn{5}{l}{\textit{GPT-6 Luna}} \\
SiLVR~\cite{zhang2026silvr} & \textbf{29.5} & 40.0 & 37.0 & 36.1 \\
MERIT~\cite{choi2026merit} & 25.6 & 37.6 & 27.5 & 29.1 \\
\oursrowS
\modelS (ours) & 24.0 & 42.1 & \textbf{39.3} & \textbf{36.9} \\
\oursrowE
\modelE (ours) & 23.5 & \textbf{42.2} & 35.5 & 34.5 \\
\midrule
\multicolumn{5}{l}{\textit{GPT-6.1 Sol}} \\
\oursrowS
\modelS (ours) & 37.6 & \textbf{50.2} & \textbf{48.7} & \textbf{46.8} \\
\oursrowE
\modelE (ours) & \textbf{38.7} & 48.3 & 43.7 & 43.6 \\
\bottomrule
\end{tabular}
\end{table}

\noindent\textbf{Results by split on MM-Lifelong.}
\label{sec:supp_mml_split}
In \cref{tab:supp_mml_split}, we report the results on the day, week, and month splits of MM-Lifelong.
First, with GPT-5, \modelS outperforms MERIT on all three splits, by $0.3\%$, $7.7\%$, and $10.5\%$ on the day, week, and month splits, respectively.
With GPT-6 Luna, it outperforms MERIT by $4.5\%$ and $11.8\%$ on the week and month splits.
Second, with GPT-5, \modelE outperforms MERIT by $4.7\%$ and $6.0\%$ on the week and month splits, and the two achieve similar scores on the day split ($15.6\%$ vs.\ $16.0\%$).
This shows that \modelE is more effective when the videos are longer.
Third, with GPT-6 Luna, \modelS outperforms SiLVR by $2.1\%$ and $2.3\%$ on the week and month splits.
SiLVR is also much more costly, since its LLM processes the captions of the entire video for every question, which leads to the highest API cost among all methods in \cref{tab:efficiency}.
Lastly, \modelE is within $1.1\%$ of \modelS on the day split with every LLM, and the difference between the two is the largest on the month split ($3.8\%$ to $5.0\%$).

\section{Qualitative Results}
\label{sec:supp_qual}

\begin{figure*}[t]
\centering
\includegraphics[width=\linewidth]{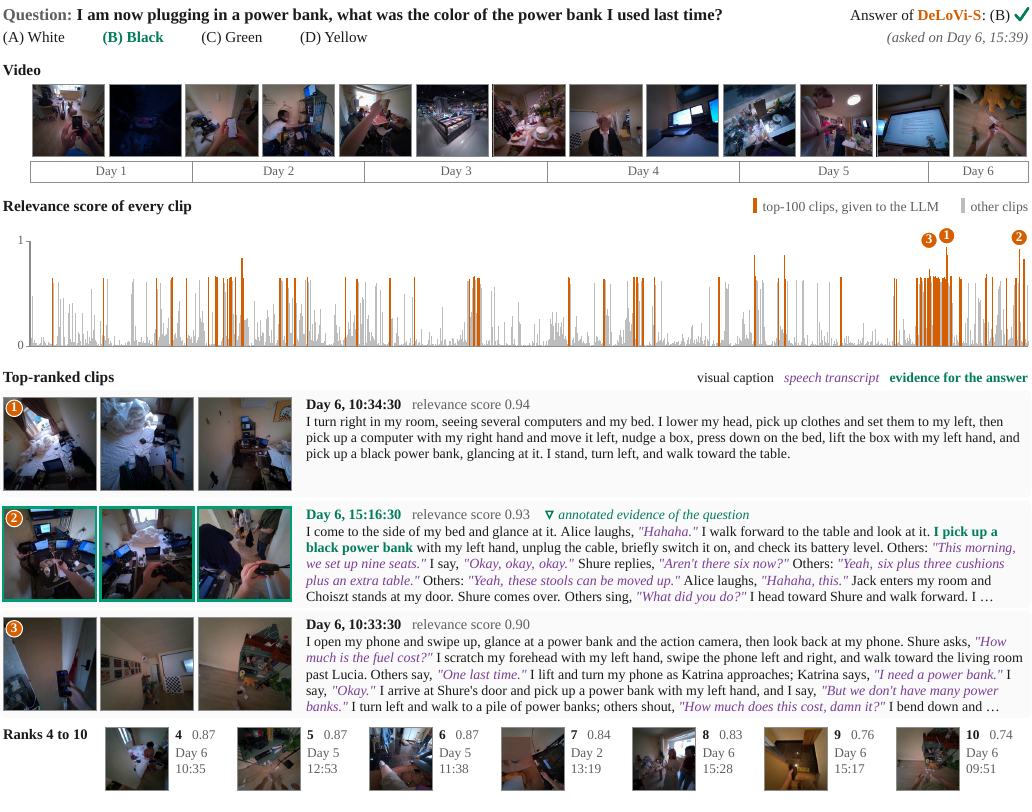}
\caption{\textbf{Qualitative example of \modelS.}
The question asks for the color of the power bank that the camera wearer used last time and is asked on Day~6 of EgoLife, after $41$ hours of video.
The decision model assigns a relevance score to the memory entry of every clip, and the $100$ clips with the highest scores are given to the LLM.
We show the three top-ranked clips with their frames and memory entries.
All three clips show the camera wearer handling a power bank on Day~6.
The top-ranked clip is from the morning of Day~6 and does not show the power bank used last time, and the annotated evidence is ranked second.
The LLM reads the retrieved memory entries in temporal order with their timestamps and answers correctly.}
\label{fig:supp_qual_delovi_s}
\end{figure*}

\begin{figure*}[t]
\centering
\includegraphics[width=\linewidth]{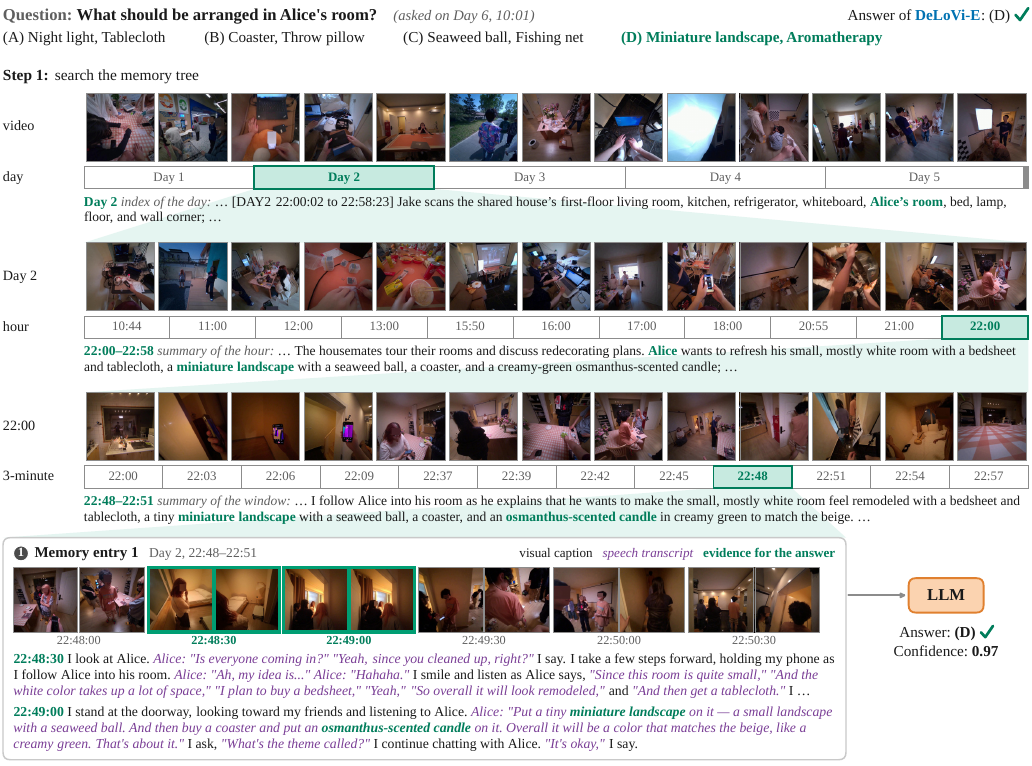}
\caption{\textbf{Qualitative example of \modelE: answering from the first evidence window.}
The question asks what should be arranged in Alice's room and is asked on Day~6, after $37$ hours of video.
In Step~1, the decision model selects Day~2, then the hour from 22:00 to 22:58, and lastly the 3-minute window from 22:48 to 22:51 (green), based on the summaries of these nodes.
The clips of this window form the first evidence window, in which Alice describes the plans for the room.
Each numbered box shows the frames and the memory entries of one evidence window.
The LLM answers from this evidence window with a confidence of $0.97$, so the search stops after a single LLM call.}
\label{fig:supp_qual_delovi_e}
\end{figure*}

\begin{figure*}[t]
\centering
\includegraphics[width=\linewidth]{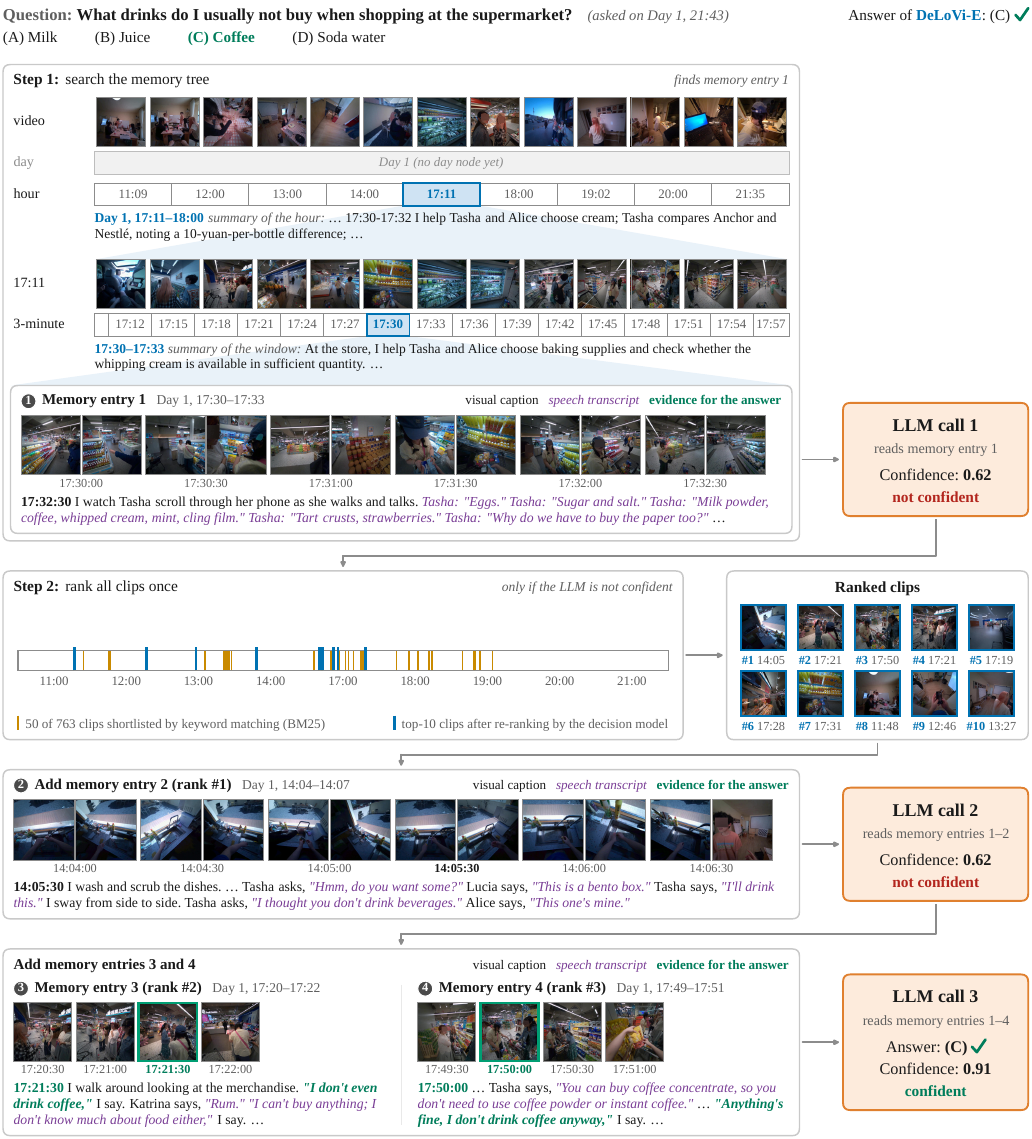}
\caption{\textbf{Qualitative example of \modelE: clip re-ranking after an answer that is not confident.}
The question asks which drinks the camera wearer usually does not buy at the supermarket.
In Step~1, the decision model selects the hour of the supermarket visit and one of its 3-minute windows, and the LLM is not confident in its answer from this evidence window (confidence $0.62$).
In Step~2, BM25 shortlists $50$ of the $763$ clips, and the decision model re-ranks them.
We show the ten top-ranked clips.
Each of the first three forms a further evidence window together with its neighboring clips (boxes~2 to~4).
The third and fourth evidence windows contain two statements of the camera wearer about not drinking coffee.
The LLM answers after the 1st, 2nd, and 4th evidence windows, and its confidence reaches $0.91$ after the 4th, so the search stops after three LLM calls.}
\label{fig:supp_qual_delovi_e_multi}
\end{figure*}

\noindent\textbf{Qualitative example of \modelS.}
In \cref{fig:supp_qual_delovi_s}, we show an example of \modelS on EgoLifeQA.
The question is asked on Day~6 and asks for the color of the power bank that the camera wearer used last time.
The video before the question contains $41$ hours of video.
The decision model scores the memory entry of every clip.
The three top-ranked clips all show the camera wearer handling a power bank on Day~6.
The top-ranked clip, with a score of $0.94$, shows the camera wearer picking up a black power bank in the morning of Day~6.
This clip matches the content of the question, but it is not the annotated evidence, since the question asks for the power bank used last time and the camera wearer uses a power bank again later on the same day.
The annotated evidence is the clip ranked second, with a score of $0.93$, in which the camera wearer picks up a black power bank $23$ minutes before the question.
The LLM reads the $100$ retrieved memory entries in temporal order, each with its timestamp, and answers the question correctly.
This suggests that \modelS does not rely on the evidence being ranked first, since the LLM can compare the retrieved moments by their timestamps.

\noindent\textbf{Qualitative examples of \modelE.}
\cref{fig:supp_qual_delovi_e,fig:supp_qual_delovi_e_multi} show two examples of \modelE on EgoLifeQA, which require one and three LLM calls, respectively.
In \cref{fig:supp_qual_delovi_e}, the question asks what should be arranged in Alice's room and is asked on Day~6, after $37$ hours of video.
In Step~1, $\mathcal{D}$ first selects Day~2, whose index mentions Alice's room in the line of its last hour.
It then selects this hour, whose summary states that Alice wants to decorate the room with a miniature landscape and a scented candle, and lastly the 3-minute window from 22:48 to 22:51.
The clips of this window form the first evidence window, in which Alice describes these plans.
The LLM answers from this evidence window with a confidence of $0.97$, which exceeds the threshold of $0.8$, so the search stops after a single LLM call.
Although the evidence lies more than three days before the question, $\mathcal{D}$ reaches it by ranking only the summaries, instead of scoring all fine-grained memory entries.

In \cref{fig:supp_qual_delovi_e_multi}, the question asks which drinks the camera wearer usually does not buy at the supermarket.
It is asked on the evening of Day~1, when the memory tree does not contain a day node yet, so the hierarchical search starts from the hours.
$\mathcal{D}$ selects the hour of the supermarket visit and the 3-minute window from 17:30 to 17:33, in which the group chooses baking supplies.
This evidence window contains no statement about the drinks that the camera wearer does not buy, and the LLM is not confident in its answer (confidence $0.62$).
\modelE therefore re-ranks the clips of the entire video in Step~2, where BM25 shortlists $50$ of the $763$ clips and $\mathcal{D}$ ranks them.
The second evidence window shows a conversation about beverages while the camera wearer washes the dishes, and the confidence remains at $0.62$.
The third and fourth evidence windows are two other moments of the supermarket visit, in which the camera wearer says ``I don't even drink coffee'' and ``I don't drink coffee anyway''.
With these evidence windows, the LLM answers correctly with a confidence of $0.91$, and the search stops after three LLM calls.
Together, the two examples show that \modelE stops after the first evidence window when the hierarchical search finds the evidence, and that the clip re-ranking retrieves evidence from other parts of the video when the LLM is not confident.

\ifdefined\supponly
\clearpage
{
    \small
    \bibliographystyle{ieeenat_fullname}
    \bibliography{main}
}
\fi

\end{document}